\documentclass[final]{cvpr}

\usepackage{times}
\usepackage{epsfig}
\usepackage{graphicx}
\usepackage{amsmath}
\usepackage{amssymb}

\usepackage[dvipsnames, table]{xcolor}
\usepackage{enumitem}

\usepackage[font=footnotesize,labelfont=bf]{caption}
\usepackage{array}
\usepackage{multirow}
\usepackage{booktabs}
\usepackage{makecell}
\usepackage{color, colortbl}

\usepackage{xspace} % for abbreviation macros

\usepackage[pagebackref=true,breaklinks=true,colorlinks,citecolor=ForestGreen,bookmarks=false]{hyperref}

\usepackage{cleveref}
\crefformat{section}{\S#2#1#3}
\crefformat{subsection}{\S#2#1#3}
\crefformat{subsubsection}{\S#2#1#3}
\crefformat{figure}{Figure~#2#1#3}
\crefformat{equation}{Equation~#2#1#3}
\crefformat{table}{Table~#2#1#3}
\crefformat{algorithm}{Algorithm~#2#1#3}

\renewcommand{\a}{\mathbf{a}}
\renewcommand{\v}{\mathbf{v}}

\newcommand{\s}{\mathbf{s}}
\renewcommand{\t}{\mathbf{t}}

\newcommand{\abar}{\bar{\a}}
\newcommand{\vbar}{\bar{\v}}

\newcommand{\tbar}{\bar{\t}}

\newcommand{\Lcal}{\mathcal{L}}
\newcommand{\Dcal}{\mathcal{D}}

\newlength\savewidth
\newlength\thinwidth

\usepackage{pifont}% http://ctan.org/pkg/pifont
\newcommand{\cmark}{\ding{51}}%
\newcommand{\xmark}{\ding{55}}%

\makeatletter
\DeclareRobustCommand\onedot{\futurelet\@let@token\@onedot}
\def\@onedot{\ifx\@let@token.\else.\null\fi\xspace}

\def\eg{\emph{e.g}\onedot} 
\def\ie{\emph{i.e}\onedot} 
 
 \def\vs{\emph{vs}\onedot}

\makeatother

\usepackage{microtype}

\renewcommand{\paragraph}[1]{\noindent {\bf #1}}

\def\cvprPaperID{2616} % *** Enter the CVPR Paper ID here
\def\confYear{CVPR 2021}

\begin{document}

%%%%%%%%% TITLE
\title{Robust Audio-Visual Instance Discrimination}

\author{Pedro Morgado\thanks{Contacting author.}\\
UC San Diego 
\and
Ishan Misra\\
Facebook AI Research
\and
Nuno Vasconcelos\\
UC San Diego
}

\maketitle
\pagestyle{empty}
\thispagestyle{empty}

%%%%%%%%% ABSTRACT
\begin{abstract}
We present a self-supervised learning method to learn audio and video representations. Prior work uses the natural correspondence between audio and video to define a standard cross-modal instance discrimination task, where a model is trained to match representations from the two modalities. However, the standard approach introduces two sources of training noise. First, audio-visual correspondences often produce faulty positives since the audio and video signals can be uninformative of each other. To limit the detrimental impact of faulty positives, we optimize a weighted contrastive learning loss, which down-weighs their contribution to the overall loss. Second, since self-supervised contrastive learning relies on random sampling of negative instances, instances that are semantically similar to the base instance can be used as faulty negatives. To alleviate the impact of faulty negatives, we propose to optimize an instance discrimination loss with a soft target distribution that estimates relationships between instances. We validate our contributions through extensive experiments on action recognition tasks and show that they address the problems of audio-visual instance discrimination and improve transfer learning performance.
\end{abstract}

%%%%%%%%% BODY TEXT
\section{Introduction}
Self-supervised representation learning aims to learn feature representations that can transfer to downstream tasks without costly human annotations.
Many recent self-supervised methods~\cite{swav, moco, pirl, simclr, cld, cmc} use a variant of the instance discrimination framework~\cite{instance,exemplar}, which matches features from multiple views/augmentations of the \textit{same} instance, while distinguishing these features from those of other instances. This often relies on a contrastive loss~\cite{hadsell2006dimensionality}, where different augmentations are considered `positives' and other samples `negatives.'

Cross-modal instance discrimination (xID) extends instance discrimination to the realm of multiple modalities, where data modalities, such as video, audio, or text, act as the different `views' of an instance.
Since there is a strong correlation between audio and visual events (\eg, the sound of an instrument or a baseball match), audio-visual instance discrimination has gained popularity~\cite{l3, multisensory, bruno_avts, avid, evolving_losses, xdc, gdt, versatile_nets}. Representations learned by these methods show promising performance on tasks like action recognition and environmental sound classification. 
xID methods rely on two key assumptions - (1) the audio and video of a sample are informative of each other, \ie, positives; (2) the audio and video of all other samples are not related, \ie, negatives. In practice, both these assumptions are too strong and do not hold for a significant amount of real-world data. This results in \emph{faulty positive} samples that are not related to each other and \emph{faulty negative} samples that are semantically related.

\begin{figure}[t!]
    \centering
    \includegraphics[width=\linewidth]{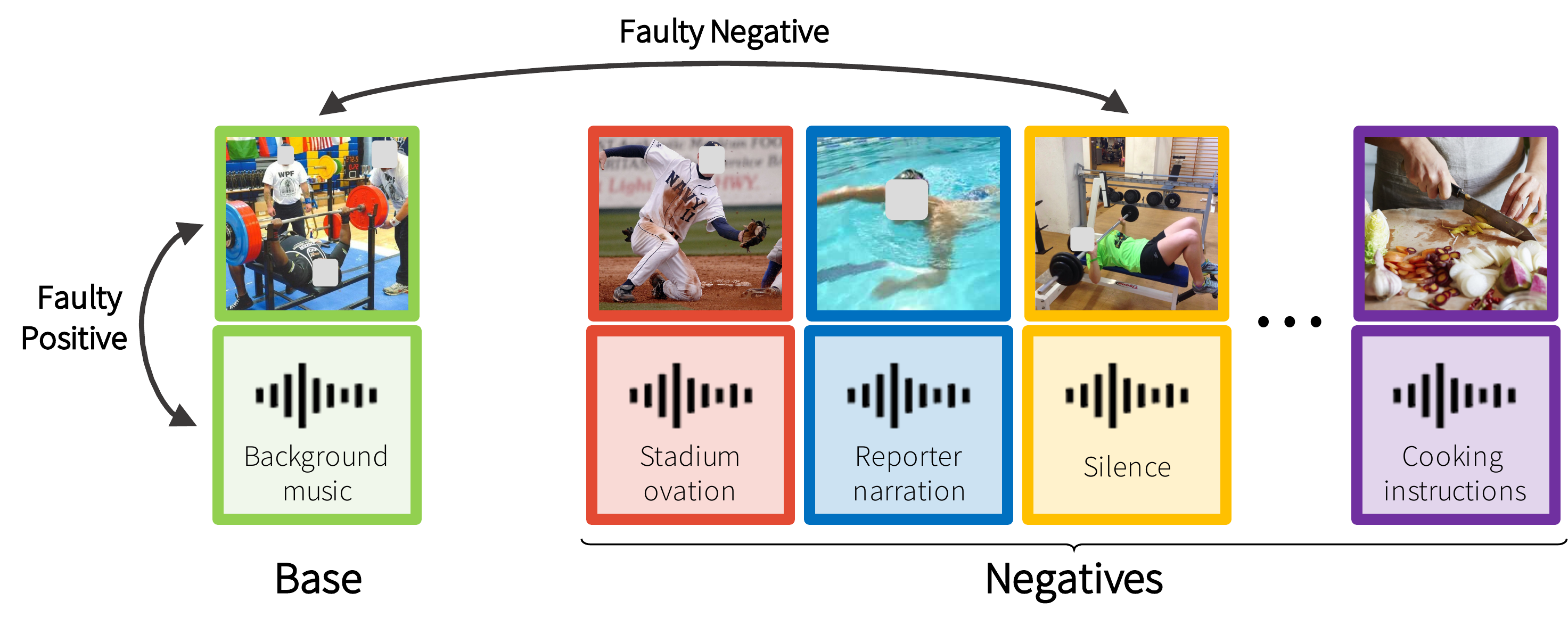}
    \caption{Example of a positive audio/video pair and negative instances used for contrastive learning. 
    Audio-visual signals may not semantically correspond to each other, such as the pairing weightlifting video/background music shown in green, which leads to faulty positives.
    Due to random sampling, semantically similar instances can also appear as faulty negatives, \eg a second weightlifting video in yellow. Faulty positive and negative samples are a common occurrence in audio-visual contrastive learning and hurt representation learning.
    }
    \label{fig:teaser}
\end{figure}

\begin{figure*}[t!]
    \centering
    \resizebox{0.94\linewidth}{!}{
    \begin{tabular}{c|c}
    \includegraphics[width=0.41\linewidth]{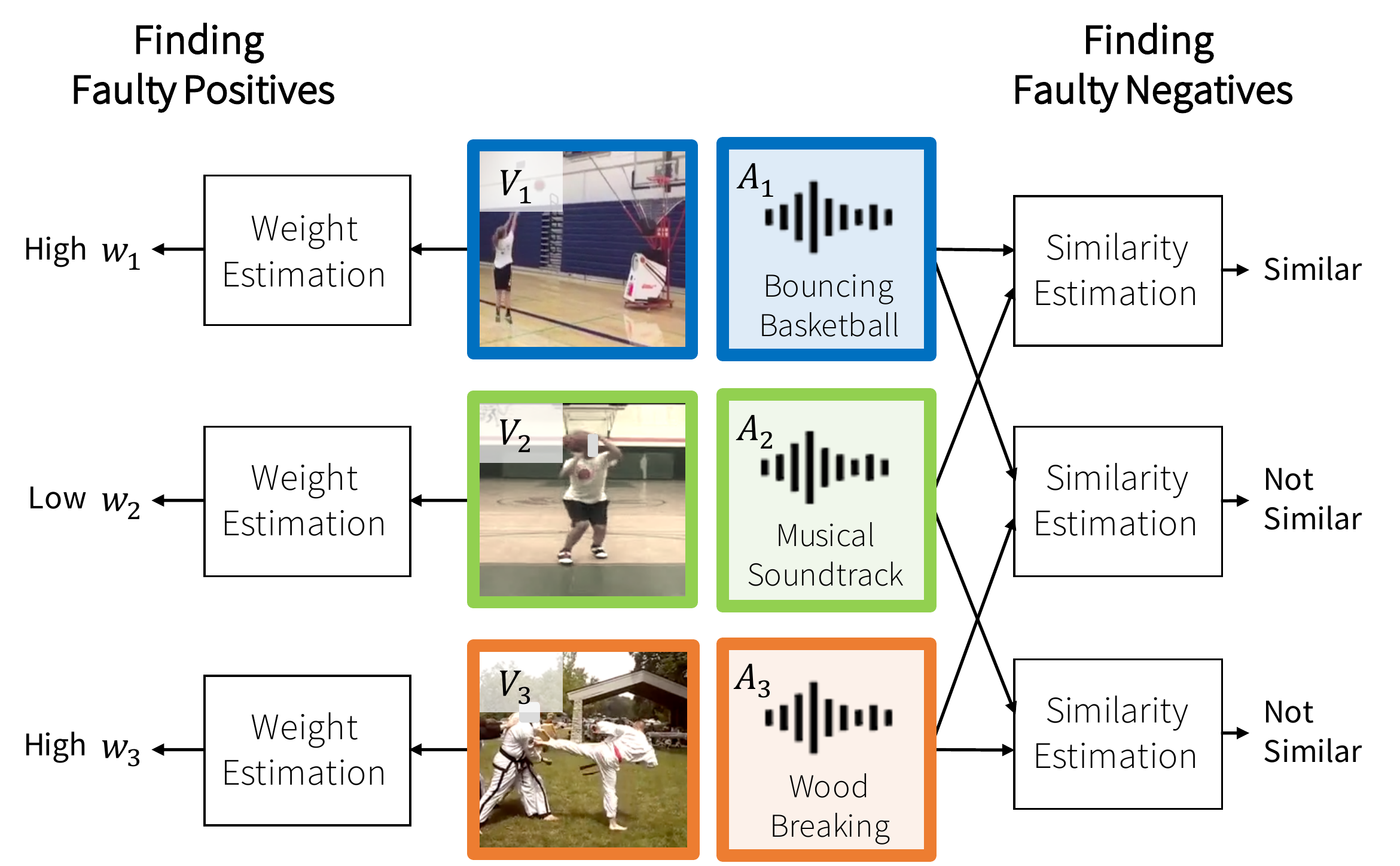} &
    \includegraphics[width=0.57\linewidth]{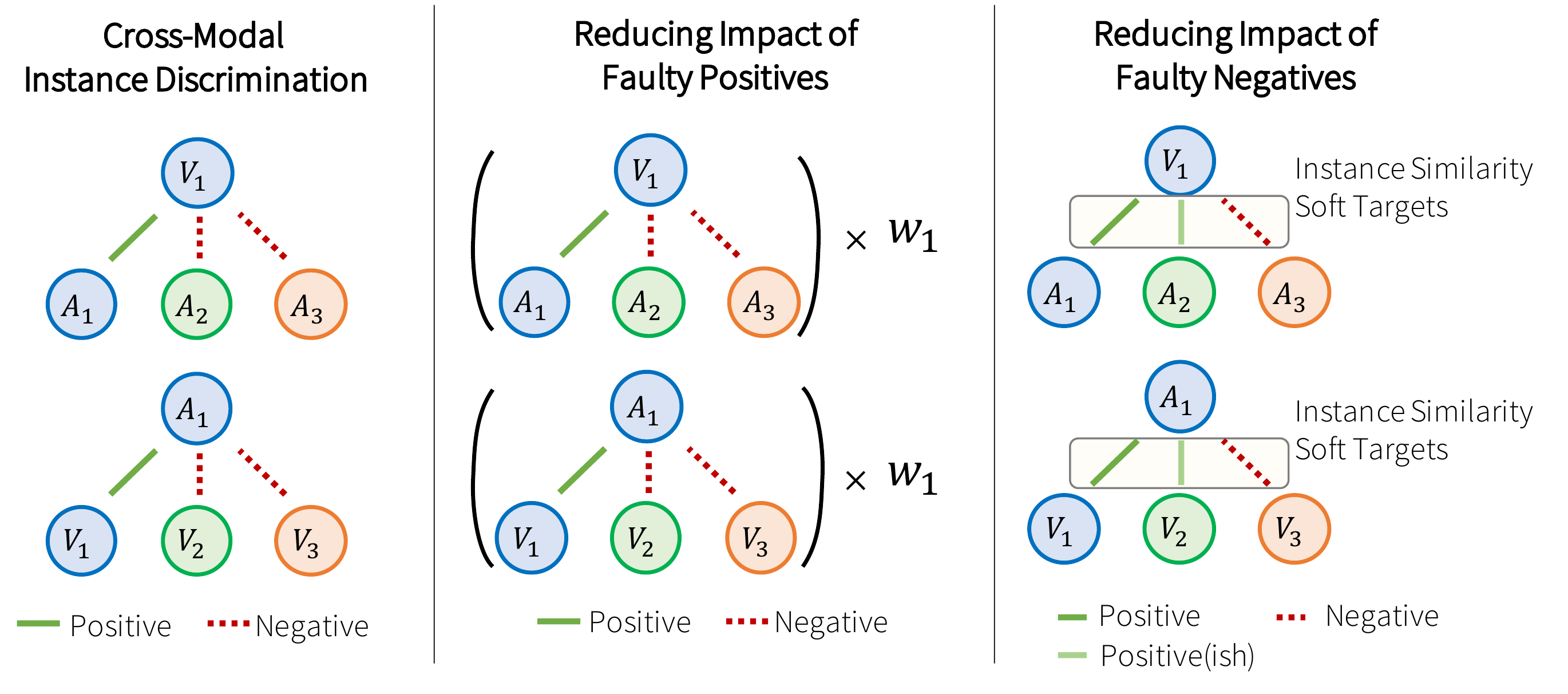}
    \end{tabular}
    }
    \caption{Comparison between standard cross-modal instance discrimination (xID) and the proposed procedure. In xID, samples contribute equally to the loss, and all instances other than themselves are treated as negatives. The proposed method addresses the two main sources of noisy training signals: faulty positives and faulty negatives. Faulty positives are discounted by down-weighting instances with poor audio-visual correspondence. Faulty negatives are addressed by optimizing the loss over a soft target distribution that encodes instance similarity.}
    \label{fig:overview}
\end{figure*}

\cref{fig:teaser} shows examples of these faulty correspondences. Videos where the audio is uninformative of the visual content can lead to faulty positives, \eg,~videos containing audio from sources outside of the camera field-of-view or containing post-edited sounds like a soundtrack. Similarly, random negative sampling can produce faulty negatives, \ie, negative samples that are semantically related to the positive. These faulty correspondences undermine the primary goal of representation learning, \ie, to ensure that similar instances have similar feature representations. As we show empirically in \cref{fig:noisy} and \cref{tab:distill-study}, they can hurt representation learning and degrade downstream performance. Thus, we believe cross-modal learning should be seen as a problem of learning with \emph{noisy targets}. This raises the question of how to identify faulty positive and negative samples in the absence of human annotations. 

We propose to use cross-modal information during self-supervised training to detect both faulty positive and negative instances. This is done by estimating the quality of the audio-visual correspondence of each instance and optimizing a weighted contrastive learning loss that down-weighs the contribution of faulty positive examples. To address faulty negatives, we estimate the similarity \emph{across} instances to compute a soft target distribution over instances. The model is then tasked to match this distribution. As a result, instances with enough evidence of similarity are no longer used as negatives and may even be used as positives. 

The contributions of this work are as follows (\cref{fig:overview}). We identify two sources of training noise in cross-modal learning: instances with weak cross-modal correspondence, which create \emph{faulty positives}, and the sampling of semantically similar instances as negatives, which create \emph{faulty negatives}.
We show that removing faulty positives and negatives using an oracle can significantly improve the performance of a state-of-the-art xID method~\cite{avid}.
We then propose a mechanism to replace the oracle and a robust cross-modal instance discrimination loss that limits the impact of faulty correspondences. The effectiveness of the proposed method is demonstrated on several downstream tasks.

\section{Related work}

\par \noindent \textbf{Self-supervised representation learning} aims to learn representations by solving pretext tasks defined from the data alone, \ie without human annotations.
In computer vision, pretext tasks involve reasoning about spatial context~\cite{jigsaw, doersch2015unsupervised, 3d_puzzle, inpainting, rotation, MemDPC, cvrl}, temporal context~\cite{shuffle, opn, aot, 3d_puzzle, mobahi2009deep, dpc, OddOneOut, pace_pred, speednet, MemDPC, CoCLR, cvrl}, other visual properties such as hue, brightness and flow~\cite{deshpande2015colorization, larsson2016colorization, colorization, larsson2017colorization, splitbrain, cmc, evolving_losses}, or clusters of features~\cite{deepcluster,sela,swav,cld}.
One promising technique is the instance discrimination task proposed in~\cite{instance, exemplar} and further explored in~\cite{moco, pirl, simclr, cld, xie2020delving}. 
However, contrastive learning from a single modality requires heavy data augmentations to generate distinct views.
Instead, we focus on cross-modal instance discrimination, which avoids this issue by generating views from different modalities.

\paragraph{Representation learning from audio-visual correspondences:}
Since, in video, the audio is naturally paired and synced with the visual component, audio-visual correspondences have been used to draw direct supervision for several tasks, such as visually guided-source separation and localization~\cite{gao2018learning, gao2019co, zhao2018sound, zhao2019sound, gan2020music, senocak2018learning}, visually guided audio spatialization~\cite{morgado2018self, 25DSound}, audio-visual embodied navigation~\cite{chen2019audio}, lip-speech synchronization~\cite{chung2016out} and audio-visual speech recognition~\cite{afouras2018deep, chung2017lip}.

In the context of contrastive learning, audio-visual correspondences are used to generate alternative views of an instance. 
While this has been known for a long time~\cite{desa94}, self-supervised audio-visual representation learning gained popularity in recent years.
For example, \cite{l3, arandjelovic2018objects} propose to learn representations by solving a \textit{binary} classification problem that identifies audio and video clips belonging to the same instance. \cite{bruno_avts, multisensory} predict if audio and video clips are temporally synchronized, and \cite{morgado20_avsa} predicts if audio and video clips extracted from a 360 video are spatially aligned. \cite{avid, gdt} improve upon the audio-visual correspondence problem~\cite{l3} by posing it as a cross-modal instance discrimination task, where instances are contrasted to a large number of negatives. As a result, \cite{avid, gdt} achieve impressive performance on downstream tasks such as action recognition.

In this work, we address two issues inherent to cross-modal instance discrimination, namely the detrimental impact of faulty positives and negatives.
Recently, \cite{xdc, selavi} proposed to learn representations by iteratively clustering the audio and visual representations and seeking to predict cluster assignments from the opposite modality. While clustering can also discourage faulty negatives from acting as repelling forces, our method accomplishes this by optimizing a simple instance discrimination loss with soft targets, thus avoiding the significant computational overhead of clustering.

\paragraph{Supervised learning from noisy labels.}
Our work is closely related to supervised learning from noisy labels~\cite{bootstrapping,GenXEnt,patrini2017making,han2018co,li2017learning}. 
Since label collection is expensive and time-consuming, scaling human annotation to large datasets often requires the use of non-experts or non-curated labels such as user tags, which are prone to noise. Since deep neural networks can easily overfit to noisy labels~\cite{Zhang2017Understanding}, this results in poor generalization. Several techniques have been developed to increase the robustness of learning algorithms to label noise, including losses that reduce the impact of outliers~\cite{rmae,GenXEnt,wang2019symmetric}, loss correction approaches that model the sources of label noise~\cite{patrini2017making,hendrycks2018using,chang2017active,bootstrapping,arazo2019unsupervised,ma2018dimensionality,song2019selfie}, meta-learning procedures that learn how to correct the sources of label noise~\cite{li2017learning,ren2018learning,li2019learning,shu2019meta,zhang2020distilling} and regularization procedures tailored to lower the impact of noise~\cite{mixup,smoothing}.
We refer the reader to \cite{song2020learning, frenay2013classification} for a detailed survey of prior work on learning with label noise.
In this work, we show that cross-modal instance discrimination should be seen as a problem of learning with noisy targets. 
However, instead of the class mislabeling, we identify two main sources of noise for cross-modal instance discrimination (faulty positives and faulty negatives) and propose an algorithm to mitigate them.

\section{Analysis: Instance Discrimination}
\label{sec:analysis_instance}

We analyze the cross-modal instance discrimination method~\cite{avid,cmc,gdt} and show that faulty positives and negatives have a disproportionately large contribution to the training updates. Additionally, in~\cref{tab:distill-study}, we document the detrimental empirical effects of faulty samples.

\paragraph{Cross-Modal Instance Discrimination}
\noindent Consider a dataset $\Dcal=\{(v_i, a_i)_{i=1}^N\}$ containing $N$ samples (or instances) of video $v_i$ and audio $a_i$. 
Cross-modal instance discrimination uses a contrastive loss~\cite{hadsell2006dimensionality} to learn video and audio encoders, $f_v(\cdot)$ and $f_a(\cdot)$, so as to align the two modalities belonging to the same instance~\cite{cmc,avid,gdt} by minimizing 
\begin{align}
    \label{eq:avid} 
    L_{\mbox{xID}}(\v_i, \a_i) =& - \log P(\abar_i | \v_i; \tau) - \log P(\vbar_i | \a_i; \tau) 
     \\
    \label{eq:avid2}
    \mbox{where \quad} & P(\tbar_i|\s_i; \tau) = \dfrac{\exp(\s_i^T\tbar_i / \tau)}{\sum_k \exp(\s_i^T\tbar_k/\tau)},
\end{align}
where $\v_i=f_v(v_i)$ and $\a_i=f_a(a_i)$ are visual and audio features normalized to the unit sphere, $\vbar_i$ and $\abar_i$ are target representations, and $\tau$ is a temperature hyper-parameter. 
Prior works differ by the type of target representations employed.
For example, $\vbar_i$ and $\abar_i$ can be entries of a memory bank as in~\cite{avid, instance}, the network representations themselves $\vbar_i=f_v(v_i)$ and $\abar_i=f_a(a_i)$ as in SimCLR~\cite{simclr}, the outputs of momentum encoders as in MoCo~\cite{moco}, or the centroids of an online clustering procedure as in SwAV or CLD~\cite{swav,cld}.
In this work, we build on the Audio-Visual Instance Discrimination (AVID) method of~\cite{avid}, focusing on target representations sampled from a memory bank. However, the principles introduced below can also be applied to SimCLR, MoCo or SwAV style targets.

\paragraph{Faulty positives and negatives in practice.}
The contrastive loss of \cref{eq:avid} is minimized when audio and visual representations from the same instance are aligned (dot-product similarities $\v_i^T\abar_i$ and $\a_i^T\vbar_i$ as close to $1$ as possible), and representations from different instances are far apart.
In practice, however, the two modalities are not informative of each other for a significant number of instances (see~\cref{fig:teaser}). We refer to these unclear correspondences as \emph{faulty positives}.\footnote{We prefer `faulty positives' over `false positives' to distinguish from supervised learning where one has access to labels.} On the other hand, a significant number of contrastive learning negatives are semantically similar to the base instance.  We term these semantically similar negatives as \emph{faulty negatives} since they should ideally be used as positives.

\begin{figure}
    \centering
    \includegraphics[width=0.9\linewidth]{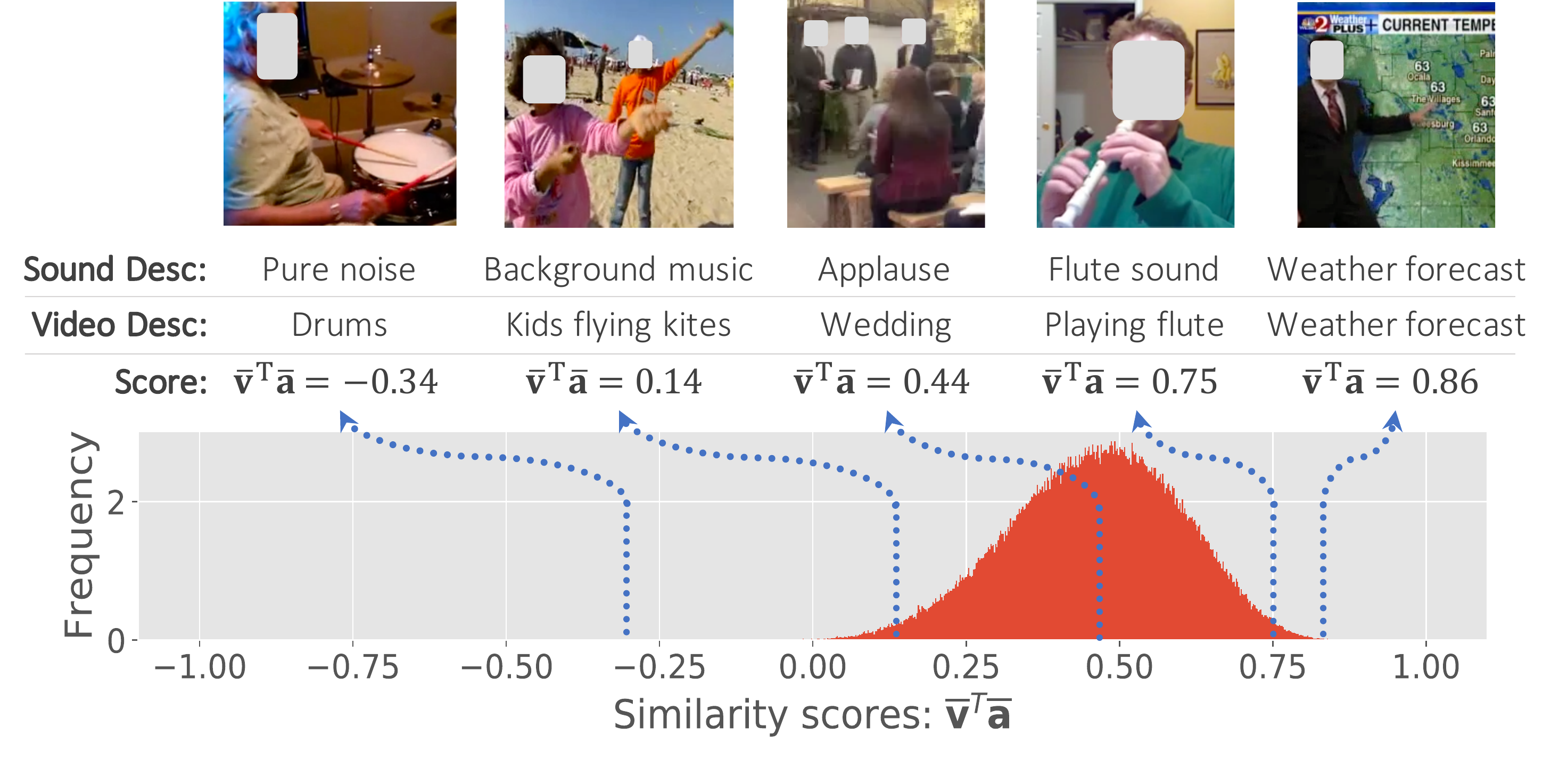}
    \caption{\textbf{Faulty positives in a pretrained cross-modal model.} Histogram of similarity scores $\vbar_i^T\abar_i$ between video and audio representations, and examples obtained at various points of the distribution. We describe both the sound and video content in the plot. Examples with lower similarity scores contain audio that is less predictive of the video content, which creates faulty positives for training. }
    \label{fig:avid_scores}
\end{figure}

\begin{figure}[t!]
    \centering
    \includegraphics[width=0.9\linewidth]{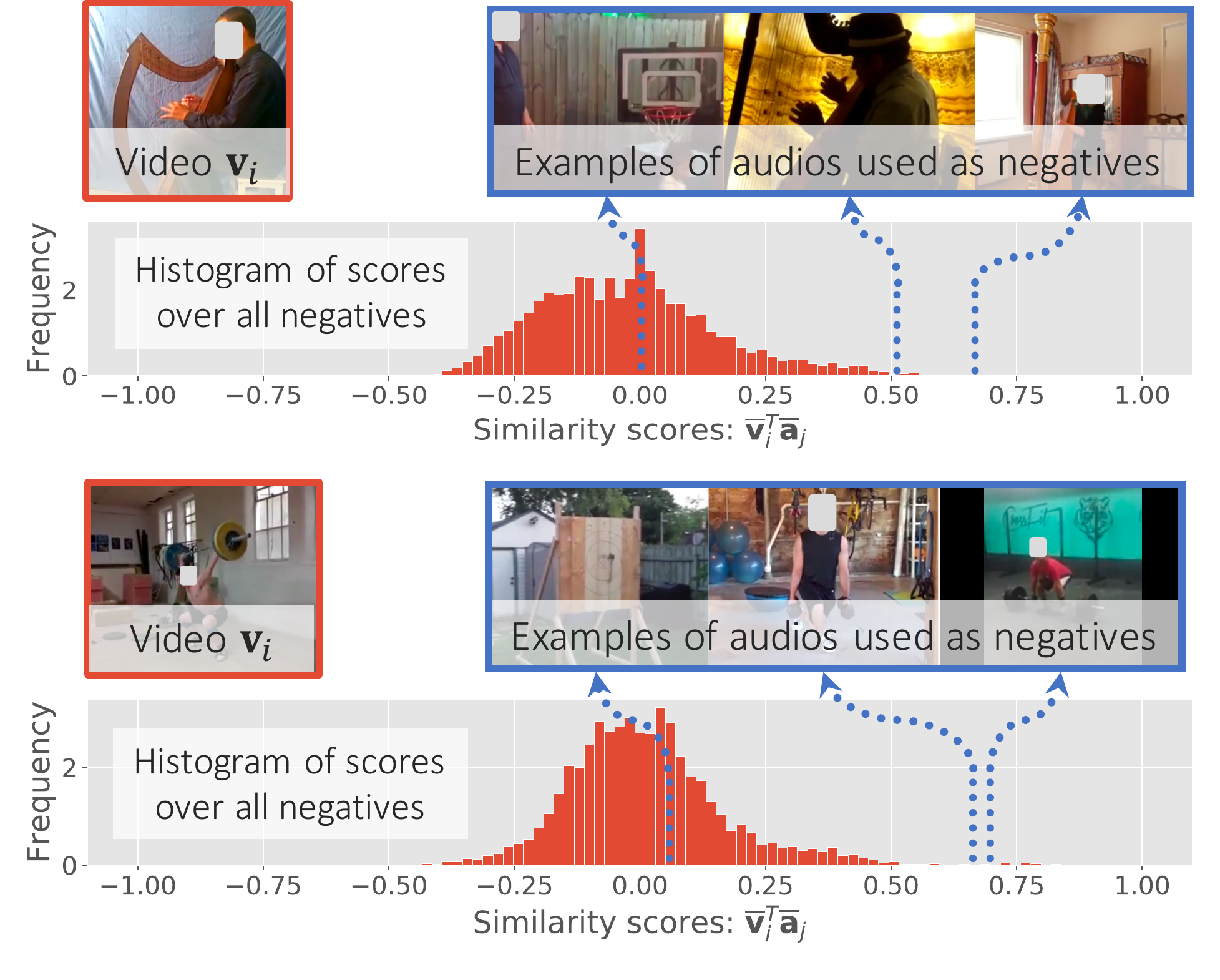}
    \caption{\textbf{Faulty negatives in a pretrained cross-modal model.} Two instances $\v_i$ and the corresponding negatives used by a xID model sorted by their similarity scores. The actual videos are provided in supplementary material. xID often uses faulty negatives for contrastive learning.}
    \label{fig:avid_negatives}
\end{figure}

\cref{fig:avid_scores} shows the histogram of similarities $\vbar_i^T\abar_i$ after training an audio-visual model with the loss of \cref{eq:avid}. As can be seen, instances with higher scores tend to have stronger correspondences (\ie the audio and video signals are informative of each other). Instances where the two modalities are uninformative of each other tend to have lower scores and are generally faulty positives.
On the other hand, \cref{fig:avid_negatives} shows the histograms of similarities between a video $i$ and negatives $j$. As can be seen, faulty negatives tend to occur for negatives $j$ with high similarity $\vbar_i^T\abar_j$.

\vspace{5pt}
\paragraph{How do faulty positives and negatives affect learning?} Faulty positives and negatives have a \emph{disproportionately large} contribution to the training updates. To see this, examine the gradients that are computed when optimizing~\cref{eq:avid}. The partial derivatives are given as
{\small
\vspace{-4pt}
\begin{align}
    -\frac{\partial L_{\mbox{xID}}}{\partial \v_i} = \underbrace{\frac{\abar_i}{\tau}(1-P(\abar_i|\v_i))}_{\text{Attraction force}}  - \underbrace{\sum_{n \neq i} \frac{\abar_n}{\tau}P(\abar_n|\v_i)}_{\text{Repulsion force}}& \label{eq:update_avid_1}\\
    -\frac{\partial L_{\mbox{xID}}}{\partial \a_i} = \underbrace{\frac{\vbar_i}{\tau}(1-P(\vbar_i|\a_i))}_{\text{Attraction force}}  - \underbrace{\sum_{n \neq i} \frac{\vbar_n}{\tau}P(\vbar_n|\a_i)}_{\text{Repulsion force}}& \label{eq:update_avid_2}.
\end{align}
\vspace{-12pt}
}

Intuitively, the target representations $\vbar_i$ and $\abar_i$ of the instance itself act as `attraction points' for the encoder of the opposing modality, while the target representations of other (negative) instances, $\vbar_n$ and $\abar_n$, act as `repelling points'. For example, in Equation~\ref{eq:update_avid_1}, the negative gradient pushes $\v_i$ toward $\abar_i$ and away from $\abar_n, n \neq i$.
The attraction forces are weighed by the complement of the prediction confidence, \ie,  $1-P(\vbar_i|\a_i)$ or $1-P(\abar_i|\v_i)$.
When positive samples are faulty, these gradients lead to noisy training signals. As show in~\cref{fig:avid_scores}, faulty positives tend to have lower similarities and thus less confident predictions. As a result, the cross-modal loss of~\cref{eq:avid} assigns stronger gradients to faulty positive samples.
On the other hand, the repelling forces of negative instances are also weighted by the likelihood of matching the base sample, \ie $P(\vbar_n|\a_i)$ and $P(\abar_n|\v_i)$. However, as shown in \cref{fig:avid_negatives}, faulty negatives tend to have high similarity scores, leading to high posteriors $P(\vbar_n|\a_i)$ and $P(\abar_n|\v_i)$. 
Thus, the targets $\vbar_n$ and $\abar_n$ of faulty negatives act as \emph{strong} repelling forces for $\a$ and $\v$ (see~\cref{eq:update_avid_1}-\ref{eq:update_avid_2}), even though they should ideally be close in feature space. 

\section{Robust audio-visual representation learning}

We have seen that contrastive learning places too much emphasis on the impossible goals of bringing together the audio-visual components of faulty positives and repelling the feature representations from faulty negatives. We next propose solutions to these two problems.

\subsection{Weighted xID: Tackling Faulty Positives}
\label{sec:rxid}

\begin{figure}[ht!]
\centering
\includegraphics[width=\linewidth]{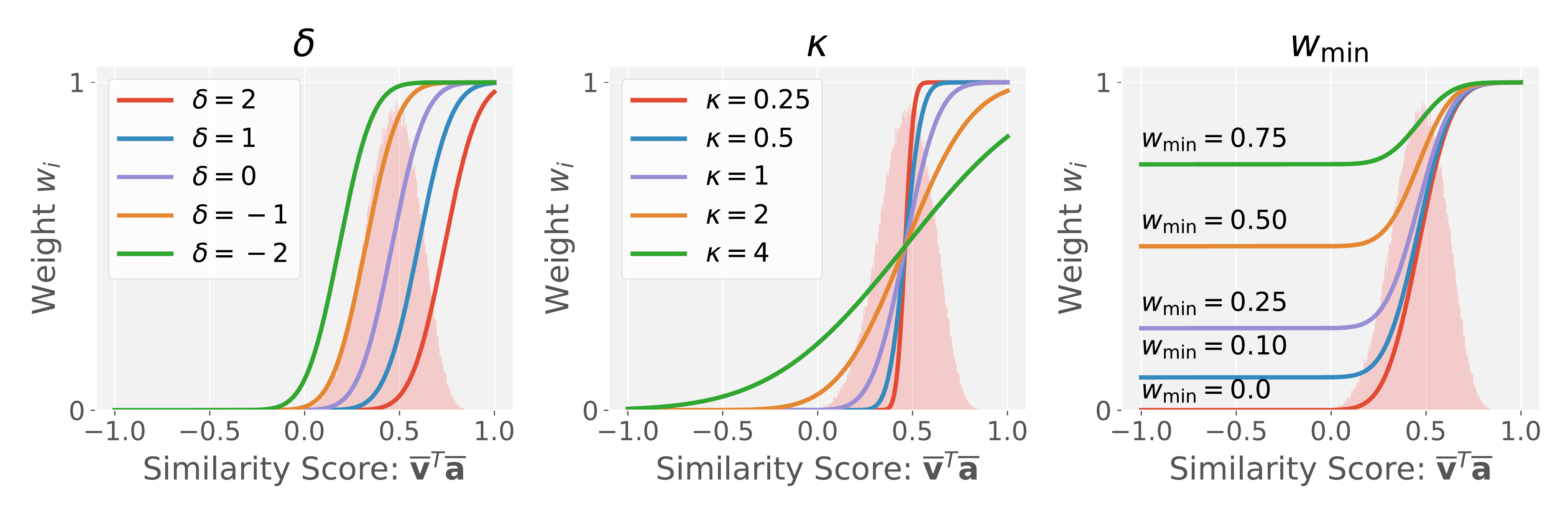}
\vspace{-15pt}
\caption{Weights as function of similarity scores $\vbar_i^T\abar_i$ for different values of shape parameters $\delta$, $\kappa$ and $w_{\min}$. Parameters $\mu, \sigma$ are automatically determined from the histogram of similarity scores $\vbar_i^T\abar_i$ (shown in red).}
\label{fig:weights}
\end{figure}

To reduce the impact of faulty positives, we propose to optimize a weighted loss. Let $w_i\in[0,1]$ be a set of sample weights that identify faulty positives.
Robustness is achieved by re-weighting the xID loss of~\cref{eq:avid}
\begin{equation}
    \Lcal_{\mbox{RxID}} = \dfrac{\sum_i w_i \Lcal_{\mbox{xID}}(\v_i, \a_i)}{\sum_i w_i}.
    \label{eq:wavid}
\end{equation}
To estimate sample weights $w_i$, we leverage observations from \cref{fig:avid_scores}. Since low similarities $\vbar_i^T\abar_i$ are indicative of faulty positives, we define the weights $w_i$ to be proportional to the cumulative distribution of these scores. We assume the scores to be normally distributed and define $w_i$ as
\begin{align}
    w_i = t_{w_{\min}}\left(C_{\mathcal{N}}\left(\abar_i^T\vbar_i; \mu+\delta\sigma, \kappa\sigma^2\right)\right),
    \label{eq:weigths}
\end{align}
where $\mu$ and $\sigma^2$ are the sample mean and variance of the scores, $C_{\mathcal{N}}$ is the cumulative distribution of a transformed normal distribution $\mathcal{N}(\mu+\delta\sigma, \kappa\sigma^2)$, and $t_{w_{\min}}(x)=x\cdot(1-w_{\min})+w_{\min}$ is a soft truncation function used to assign a non-zero weight $w_{\min}$ to low score instances.
$\delta$, $\kappa$ and $w_{\min}$ are shape hyper-parameters that provide flexibility to the weight function, adjusting the location and rate of decay of the weights.
\cref{fig:weights} shows how the weighting function varies with the shape hyper-parameters $\delta$, $\kappa$ and $w_{\min}$.

\subsection{Soft Targets: Tackling Faulty Negatives}
\label{sec:distill}

As observed in~\cref{sec:analysis_instance}, faulty negatives are overemphasized during training. The underlying reason is that the xID loss of~\cref{eq:avid} has too strict a definition of negatives: every negative instance $j\neq i$ is considered `equally negative.' To limit the impact of faulty negatives, we introduce a `softer' definition by introducing soft targets $T(j|i)$, based on the similarity between instance $i$ and negative $j$. 
We then minimize a soft-xID loss
\begin{align}
    \Lcal_{\mbox{\footnotesize Soft-xID}}(\v_i, \a_i) =
    &\textstyle - \sum_{j} T_v(j|i) \log P(\abar_j|\v_i;\tau) \nonumber \\
    &\textstyle - \sum_{j} T_a(j|i) \log P(\vbar_j|\a_i;\tau) 
    \label{eq:soft-xID} \\
    T_v(j|i)  =&\ (1-\lambda)\mathbf{1}_{i=j}+\lambda S_v(j | i) 
    \label{eq:bts-trg-v}\\
    T_a(j|i)  =&\ (1-\lambda)\mathbf{1}_{i=j}+\lambda S_a(j | i)
    \label{eq:bts-trg-a}
\end{align}
where $\mathbf{1}_{i=j}$ is the one-hot targets of vanilla xID, $S_v$ and $S_a\in[0,1]$ are softening scores (described next) used to adjust the one-hot targets, and $\lambda\in[0,1]$ is a mixing coefficient that weighs the two terms.
Equations \ref{eq:avid} and \ref{eq:soft-xID} are identical when $\lambda=0$.
Since $T(j|i)$ is no longer strictly zero for similar instances, minimizing \cref{eq:soft-xID} reduces the force to repel faulty negatives and thus their impact.

\begin{figure}[t!]
    \centering
    \includegraphics[width=0.9\linewidth]{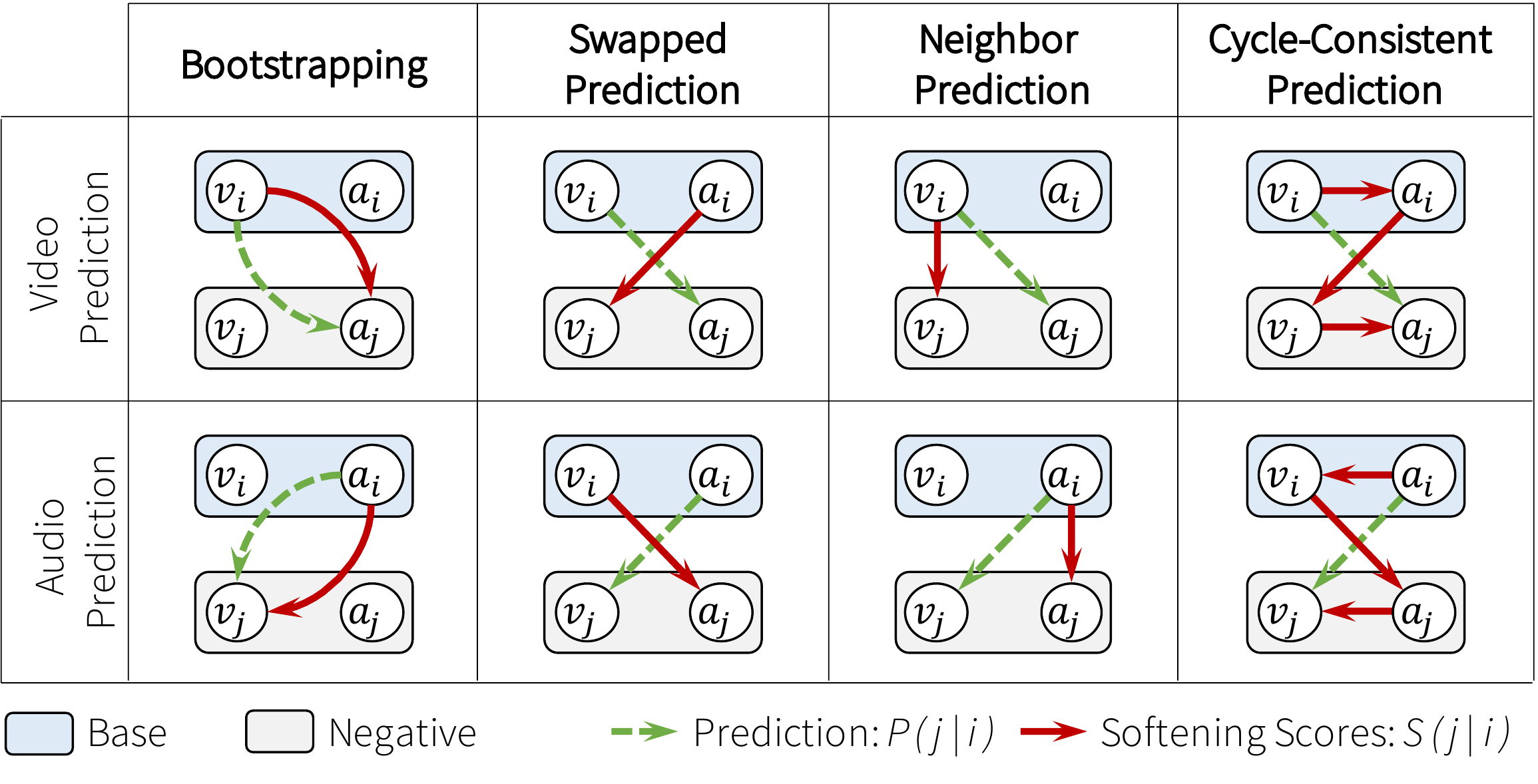}
    \caption{Strategies to estimate softening scores $S(i|j)$.}
    \label{fig:soft-targets}
\end{figure}

\vspace{5pt}
\paragraph{Estimating softening scores $S$.}
Since our approach focuses on self-supervised learning, we must estimate the softening scores $S$ automatically, \ie, without class labels. We describe multiple strategies for estimating these values and illustrate them in~\cref{fig:soft-targets}.

\vspace{5pt}\textbullet\ \ \textbf{Bootstrapping}~\cite{bootstrapping} is a well established procedure to create soft targets. It uses the model's own predictions (posteriors) as the softening scores, \ie, 
\begin{equation}
    S_v(j|i) = P(\abar_j | \vbar_i; \tau_s) \mbox{ and } S_a(j|i) = P(\vbar_j | \abar_i; \tau_s),
\end{equation} 
where $\tau_s$ controls the peakiness of the distribution. However, bootstrapping computes the target distribution without aggregating  information from any other source other than each model's own posterior.

\vspace{5pt}\textbullet\ \ \textbf{Swapped prediction} improves upon bootstrapping by using the posteriors of the opposite modality, \ie, the softening scores $S_v$ for the video modality are computed using the posterior of the audio encoder and vice-versa,
\begin{equation}
    S_v(j|i) = P(\vbar_j | \abar_i; \tau_s) \mbox{ and } S_a(j|i) = P(\abar_j | \vbar_i; \tau_s).
\end{equation}
As a result, in addition to the instance itself, the model is asked to predict which other instances are deemed similar in the opposite modality.

\vspace{5pt}\textbullet\ \ \textbf{Neighbor prediction} relies on within-modal relationships to estimate the similarity between instances, thus avoiding potential mismatched audio and visual modalities when computing the soft targets. Specifically, we define
\begin{equation}
    S_v(j|i) = \rho(\vbar_i^T\vbar_j /\tau_s) \mbox{ and } S_a(j|i) = \rho(\abar_i^T\abar_j /\tau_s),
\end{equation}
where $\rho$ is the softmax operator.

\vspace{5pt}\textbullet\ \ \textbf{Cycle consistent prediction} improves upon `swapped prediction` by focusing on negatives that are good correspondences themselves, \ie, negatives with high similarity scores $\vbar_j^T\abar_j$. In this case, we define 
\begin{align}
    S_v(j|i) = \rho(\vbar_i^T\abar_i /\tau_t + \abar_i^T\vbar_j /\tau_s  + \vbar_j^T\abar_j /\tau_t) \\ 
    S_a(j|i) = \rho(\abar_i^T\vbar_i /\tau_t + \vbar_i^T\abar_j /\tau_s  + \abar_j^T\vbar_j /\tau_t)
\end{align}
where $\tau_s$ and $\tau_t$ control the relative importance of swapped prediction target and avoiding negatives with weak correspondences. As shown in \cref{fig:soft-targets}, the terms $\vbar_i^T\abar_i$ and $\vbar_j^T\abar_j$ complete a cycle over instances $i$ and $j$.

\vspace{5pt}
\paragraph{How do soft targets mitigate faulty negatives?}
The soft xID loss of \cref{eq:soft-xID} prevents overemphasizing faulty negatives by relying on soft targets $T(j|i)$ that encode similarities between instances. To better understand the mechanism, we examine the partial derivatives of the soft-xID loss:
\begin{align}
    \textstyle -\frac{\partial L_{\mbox{\footnotesize Soft-xID}}}{\partial \v_i} =
    \sum_j \frac{\abar_j}{\tau}(T_v(j|i)-P(\abar_i|\v_i))& \label{eq:update_rxid_1}\\
    \textstyle -\frac{\partial L_{\mbox{\footnotesize Soft-xID}}}{\partial \a_i} =
    \sum_j \frac{\vbar_j}{\tau}(T_a(j|i)-P(\vbar_i|\a_i))& \label{eq:update_rxid_2}.
\end{align}
Since faulty negatives $j$ tend to be similar to the base instance $i$, the soft targets $T(j|i)$ are higher. Thus, the target representations $\vbar_j$ and $\abar_j$ of faulty negatives act as weaker negatives, or even as positives when $T(j|i)$ is larger than the model posteriors.

\subsection{Training}
We introduced two procedures to deal with noisy training signals inherent to cross-modal instance discrimination. 
\cref{sec:rxid} presents a weighting mechanism that limits the effect of faulty positives, while \cref{sec:distill} proposes a soft instance discrimination loss that predicts relations between instances, thus preventing the training algorithm from overemphasizing faulty negatives.
Since both procedures rely on the alignment between audio and visual target representations to find weak correspondences, we start by training the model for cross-modal instance discrimination alone (\cref{eq:avid}). 
After the initial warmup stage, the two procedures can be combined by minimizing
\begin{equation}
    \Lcal = \tfrac{1}{\sum_k w_k} \sum_i w_i \Lcal_{\mbox{\footnotesize Soft-xID}}(\v_i, \a_i) 
    \label{eq:loss}
\end{equation}
where $w_i$ are the sample weights of \cref{eq:weigths} and  $\Lcal_{\mbox{\footnotesize Soft-xID}}$ is the xID loss with soft targets of \cref{eq:soft-xID}.

\section{Experiments}
\label{sec:experiments}
We perform experiments to better understand cross-modal learning and validate the proposed improvements.
We pretrain models on a subset of the Kinetics-400~\cite{kinetics} dataset containing 50K videos and evaluate the pretrained models by transfer learning.

\subsection{Experimental Setup}
\par \noindent \textbf{Video and audio preprocessing.} 
During training, we extract video clips of length $T=8$ frames and resolution $80 \times 80$ at 16 fps. Video clips are augmented using temporal jittering, multi-scale cropping, horizontal flipping, color jittering, gray-scaling, and Gaussian blur~\cite{simclr}. All data augmentations are applied consistently over all frames.
For the audio, we extract mono clips of length $2$s at a sample rate of $11025$Hz, and compute log spectrograms on $50$ms windows with a hop size of $25$ms. The spectrogram is then converted to a mel scale with $80$ bands, yielding an audio input of size $80 \times 80$. Audio data is augmented by randomly changing the volume by at most $20\%$.

\par \noindent \textbf{Video and audio models.}
The video encoder is a 9-layer version of the R(2+1)D model of~\cite{tran2018closer}. Following~\cite{l3, avid}, we replaced global average pooling with max pooling. The audio encoder is a 9-layer 2D ConvNet with batch normalization and global max pooling. 
Both encoders yield 512-dimensional features, which are mapped into a 128-dimensional sphere using a non-linear projection head (as in~\cite{simclr}) followed by L2 normalization.

\par \noindent \textbf{Pretraining.}
In the warm-up stage, the video and audio models are trained to optimize the loss of~\cref{eq:avid} using the Adam optimizer~\cite{adam} with default hyper-parameters ($\beta_1=0.9$ and $\beta_2=0.999$) for $400$ epochs with a learning rate of $1e-4$ and a batch size of $224$ split over 2 $12$Gb GPUs. In order to reduce the memory footprint of our models, we employ mixed-precision training~\cite{micikevicius2017mixed} using PyTorch AMP~\cite{pytorch}.
Following~\cite{avid,instance}, the audio and video target representations, $\abar$ and $\vbar$, are generated using memory banks updated by exponential moving average with an update constant of $0.5$.
The contrastive loss of~\cref{eq:avid} is defined by opposing the target representation of the opposite modality to $1024$ negatives randomly drawn from the memory bank. The temperature hyper-parameter is set to $\tau=0.07$.

After the initial warm-up stage, models are trained for an additional 200 epochs to optimize the loss of~\cref{eq:loss} using the Adam optimizer and a cosine learning rate schedule starting at $1e-4$ and ending at $1e-5$. The hyper-parameters for the weighting function (\cref{eq:weigths}) and the soft xID loss (\cref{eq:soft-xID}) are discussed below.
To provide a fair comparison to the AVID baseline~\cite{avid}, we control for the number of epochs by training the baseline model for an additional 200 epochs as well.

\par \noindent \textbf{Downstream tasks.}
We evaluate audio and video features using transfer learning. 
Video features are evaluated on the UCF~\cite{ucf} and HMDB~\cite{hmdb} datasets. Models are fine-tuned using 8-frame clips for 200 epochs using the Adam optimizer with a batch size of $192$ on a single GPU and a cosine learning rate schedule starting at $1e-4$ and ending at $1e-5$. To prevent overfitting, we use dropout after the global max-pooling layer, weight decay of $1e-3$, and reduced the learning rate for backbone weights by a factor of $10$.
At test time, top-1 accuracy is measured on video level predictions computed by averaging the predictions of $10$ clips uniformly sampled over the entire video.

Following~\cite{selavi,xu2019self}, we also evaluate the quality of video representations by conducting retrieval experiments without fine-tuning. Feature maps of size $4\!\!\times\!\!4\!\!\times\!512$ are extracted from $10$ clips per video and averaged. We then use videos in the test set to query the training set. As in~\cite{selavi,xu2019self}, a correct retrieval occurs when the class of one of the top-k retrieved videos matches the query, and performance is measured by the average top-k retrieval performance ($R@K$).

\subsection{Weighted cross-modal learning}
We analyze the impact of faulty positives on the representations learned by cross-modal instance discrimination.

\begin{figure}
    \centering
    \includegraphics[width=0.9\linewidth]{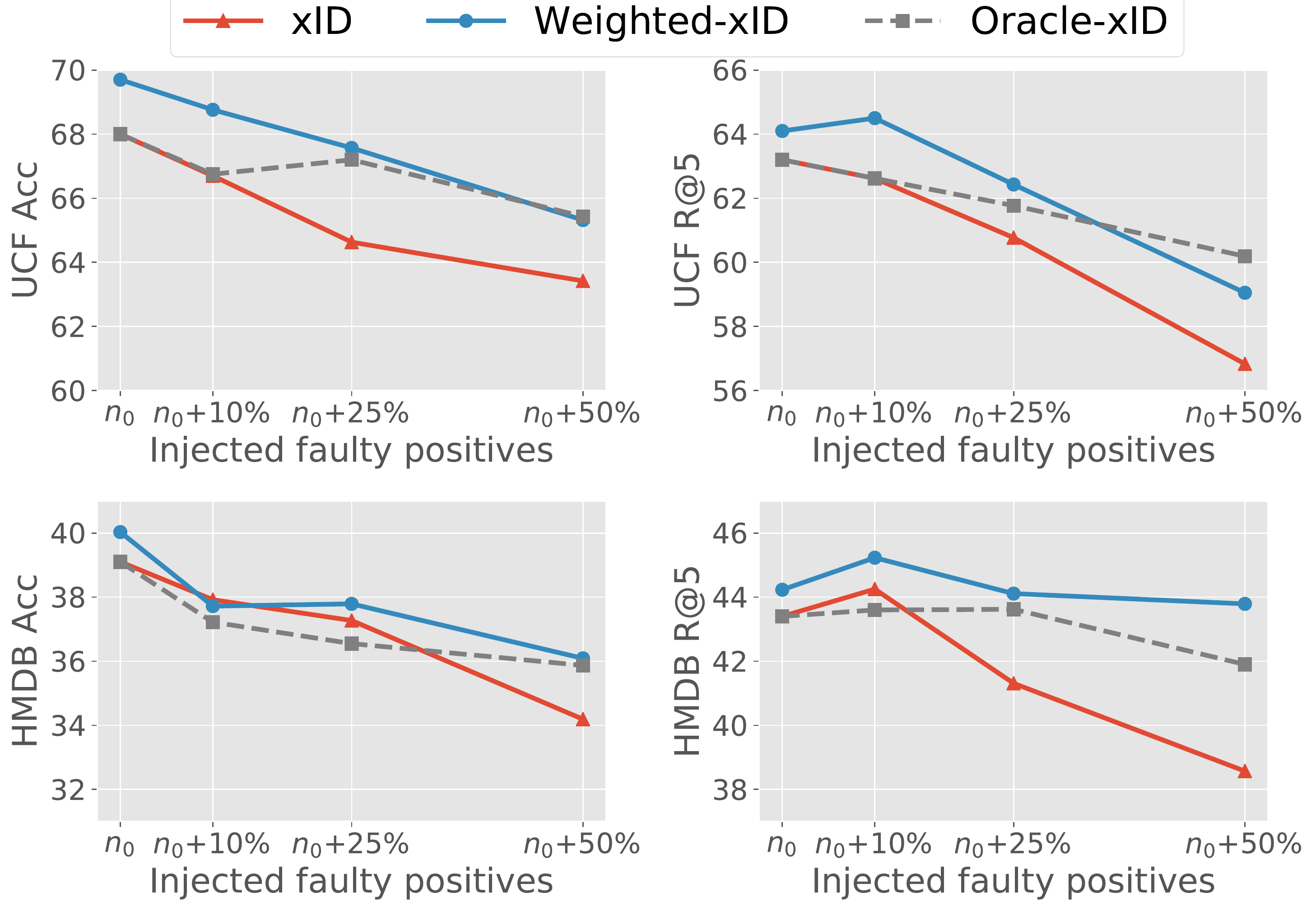}
    \caption{\textbf{Faulty positives \vs transfer performance} We inject faulty positives in the pre-training data (Kinetics) by randomly replacing the audio corresponding to a video. We evaluate the pretrained models on action recognition and see that increasing the fraction of faulty positives hurts transfer performance. Our weighted loss (Weighted-xID) is less sensitive to faulty positives and even outperforms an oracle version that information about altered samples. This is because the unaltered pretraining dataset itself has an unknown, but significant number ($n_0$) of faulty positives.
    }
    \label{fig:noisy}
\end{figure}

\paragraph{Faulty positives are detrimental to representation learning.}
We artificially control the number of faulty positives to assess their impact on representation learning. The pretraining dataset \emph{already contains} an unknown (but significant) number of faulty positives. We increase this number by injecting more faulty positives.
A faulty positive is injected by replacing the audio of an instance with a randomly selected audio that is not part of the training set.
After pretraining, the learned visual representation is evaluated on the UCF and HMDB datasets using both classification and retrieval protocols.
\cref{fig:noisy} shows that as the fraction of faulty positives increases, the transfer performance of cross-modal instance discrimination (xID) decreases significantly.

\paragraph{Weighted xID reduces the impact of faulty positives.}
We evaluate the effectiveness of the weighted xID loss (\cref{eq:wavid}) as a function of the number of faulty positives.
We compare the representations learned by Weighted-xID to its unweighted counterpart (xID), as well as an oracle weight function (Oracle-xID) which assigns $w_i=0$ to artificially altered instances and $w_i=1$ otherwise.
The weight function of~\cref{eq:wavid} is defined with $\kappa=0.5$ and $w_{\min}=0.25$. 
For simplicity, we assume that the noise level is known and set $\delta$ in Weighted-xID so that the midpoint of the weighting function coincides with the known fraction of altered samples. In practice, the noise level would need to be estimated either by cross-validation or by manual inspection. Weighted-xID is not very sensitive to these parameters (see appendix).

\cref{fig:noisy} shows the performance of the three approaches. Oracle-xID consistently outperforms xID when the fraction of injected faulty positives is high. This shows that the detrimental impact of noisy correspondences can be mitigated with a weighting strategy. Weighted-xID also outperforms the unweighted version (xID) in nearly all cases, with larger margins for larger fractions of noisy correspondences. In fact, Weighted-xID even outperforms the oracle weight function, especially at lower noise levels. This is because the original Kinetics dataset \emph{already contains} a significant amount of weak correspondences, which the oracle weight function treats as clean $w_i=1$, while the weighting function of \cref{eq:weigths} can suppress them.

\subsection{Instance discrimination with soft targets}
\label{sec:soft-xid}

To limit the impact of faulty negatives, we proposed to match a soft target distribution that encodes instance similarity. We analyze different design decisions for creating the soft targets and their effect on transfer performance.

% \begin{table}[t!]
% \centering
% \caption{Transfer learning on UCF, HMDB and ESC datasets after pre-training on Kinetics with different distillation settings. C@1 stands for top1 accuracy of clip level predictions and V@1 of video level predictions. xM: Cross-modal; wM: Within-modal; Bst: Bootstrapping; Swap: Swapped posteriors; CCP: Cycle Consistent Assignments.}
% \label{tab:distill-study}
% \resizebox{0.8\linewidth}{!}{%
% \begin{tabular}{@{}cc @{\hspace{4ex}} cc @{\hspace{4ex}} cc @{}}
% \toprule
% \multicolumn{1}{l}{\multirow{2}{*}{\thead{Targets}}} &
% \multicolumn{1}{c@{\hspace{4ex}}}{\multirow{2}{*}{\thead{Target\\Distribution}}} &
% \multicolumn{2}{c@{\hspace{4ex}}}{UCF} &
% \multicolumn{2}{c@{\hspace{4ex}}}{HMDB} \\ \cmidrule(l){3-6} 
% \multicolumn{2}{l}{} & ACC & R@5 & ACC & R@5 \\ \midrule
% \multirow{4}{*}{xM} 
% & Bst       & 69.2 & 64.4 & 40.5 & 44.7 \\
% & Swap      & 70.0 & 64.9 & 41.3 & 45.4 \\
% & CCP       & 70.3 & 65.9 & 41.5 & 45.5 \\\cline{2-6}
% & Oracle    & 73.6 & 76.0 & 45.4 & 53.6 \\
% \midrule
% \multirow{4}{*}{wM} 
% & Bst       & 69.2 & 64.1 & 39.6 & 43.4 \\
% & Swap      & 68.9 & 64.9 & 41.4 & 46.0 \\
% & CCP       & 70.0 & 66.2 & 41.9 & 46.4 \\\cline{2-6}
% & Oracle    & 72.8 & 72.4 & 44.7 & 52.0 \\ 
% \midrule\midrule
% \multicolumn{2}{c}{xID~\cite{avid}} & 68.0 & 63.2 & 39.0 & 43.4 \\ 
% \bottomrule
% \end{tabular}%
% }
% \end{table}

\begin{table}[t!]
\centering
\caption{
\textbf{Different strategies for computing soft targets} in the pretraining loss of~\cref{eq:soft-xID}. Models are pretrained on Kinetics and evaluated on UCF and HMDB datasets using fine-tuning and retrieval protocols. Best method is bolded. Second best is underlined.}
\label{tab:distill-study}
\resizebox{0.7\linewidth}{!}{%
\definecolor{rc1}{RGB}{235,235,235}
\definecolor{rc2}{RGB}{255,255,255}
\begin{tabular}{lcccccc}
\toprule
\multirow{2}{*}{\thead{\bf Target\\\bf Distribution}} &&
\multicolumn{2}{c}{\bf UCF} &&
\multicolumn{2}{c}{\bf HMDB} \\ \cmidrule{3-4} \cmidrule{6-7}
 && Acc & R@5 && Acc & R@5 \\ \midrule\midrule
\rowcolor{rc1}
Oracle${}^*$   && 73.6 & 76.0 && 45.4 & 53.6 \\
\rowcolor{rc2}
xID~\cite{avid} && 68.0 & 63.2 && 39.0 & 43.4 \\ 
\hline
\rowcolor{rc1}
Bootstrapping && 69.2 & 64.4 && 40.5 & 44.7 \\
\rowcolor{rc2}
Neighbor Pred. && \bf 70.5 & \underline{65.4} && 41.2 & 45.0 \\
\rowcolor{rc1}
Swapped Pred. && 70.0 & 64.9 && \underline{41.3} & \underline{45.4} \\
\rowcolor{rc2}
CCP       && \underline{70.3} & \bf 65.9 && \bf 41.5 & \bf 45.5 \\
\bottomrule
\multicolumn{7}{l}{\footnotesize \it ${}^*$Uses class labels to generate target distribution.}
\end{tabular}%
}
\end{table}

\paragraph{Comparison of strategies for computing targets}
As summarized in \cref{fig:soft-targets}, the soft target distributions can be computed by aggregating evidence from all modalities.
Four different strategies were proposed, bootstrapping, swapped or cycle consistent assignments.
Models were trained to minimize the loss of~\cref{eq:soft-xID} with $\lambda=0.5$. We empirically found that peakier target distributions work better, and set the temperature parameter $\tau_s$ to $0.02$. For cycle consistent assignments, the terms $\vbar_j^T\abar_j$ are used so as to focus on negatives that are good correspondences themselves. A temperature hyper-parameter of $\tau_t=0.07$ was sufficient to impose such constraint.
Beyond the baseline xID, we also compare to an \emph{oracle} target distribution that has access to class labels to determine the similarity between instances. 
Specifically, the oracle considers two instances $i$ and $j$ to be similar if they share the same class label, and computes $T_v(j|i)$ and $T_a(j|i)$ by assigning a uniform distribution over similar instances, and $0$ to non-similar ones.

\cref{tab:distill-study} shows the performance of different target distributions. We observe a large gap between vanilla xID and xID with an oracle soft target, which demonstrates the detrimental effect of faulty negatives.
In the self-supervised case, however, labels are not available for determining the target distribution.
Nevertheless, the estimated target distributions (bottom four rows) still significantly improve over the xID loss.
Regarding the various types of target distributions, bootstrapping is the least effective. This is expected since, in this case, the target distribution is a peakier version of the model posterior, \ie it is obtained without aggregating information from any other sources. Cycle consistent prediction is the most effective most often. 
This is because cycle consistent prediction not only leverages the opposite modality to create the target distribution, but it also avoids targets that are not good correspondences themselves, \ie, avoids samples with low cross-modal similarities.

\begin{table}[t!]
\centering
\caption{{\bf Combining weighted xID loss with soft targets}. Models are pretrained on Kinetics with the loss of \cref{eq:loss} and evaluated on UCF and HMDB datasets using fine-tuning and retrieval protocols.}
\label{tab:robust-distill}
\resizebox{0.9\linewidth}{!}{%
\definecolor{rc1}{RGB}{235,235,235}
\definecolor{rc2}{RGB}{255,255,255}
\begin{tabular}{lcccccccc}
\toprule
\multicolumn{1}{l}{\multirow{2}{*}{\bf Method}} &
\multicolumn{1}{l}{\multirow{2}{*}{\thead{\bf Robust\\\bf Weighting}}} &
\multicolumn{1}{l}{\multirow{2}{*}{\thead{\bf CCP\\\bf Soft Targets}}} &&
\multicolumn{2}{c}{\bf UCF} &&
\multicolumn{2}{c}{\bf HMDB} \\ \cmidrule{5-6} \cmidrule{8-9}
 & & && Acc & R@5 && Acc & R@5 \\ \midrule\midrule
\rowcolor{rc1}
xID~\cite{avid} & \xmark & \xmark && 68.0 & 63.2 && 39.0 & 43.4  \\
\rowcolor{rc2}
Weighted-xID    & \cmark & \xmark && 69.7 & 64.1 && 40.1 & 44.3 \\
\rowcolor{rc1}
Soft-xID        & \xmark & \cmark && 70.3 & 65.9 && 41.5 & 45.5 \\
\rowcolor{rc2}
Robust-xID      & \cmark & \cmark && \bf 71.6 & \bf 67.4 && \bf 41.9 & \bf 46.2 \\
\bottomrule
\end{tabular}%
}
\end{table}

\subsection{Robust instance discrimination with soft targets}
Sample weighting and soft targets are designed to address two different sources of noisy training signals inherent to cross-modal contrastive learning: faulty positives and faulty negatives.
\cref{tab:robust-distill} shows that the two proposed improvements (Weighted-xID and Soft-xID) not only improve upon the representations of vanilla xID, they are also complementary to each other.
By combining the two approaches using the loss of \cref{eq:loss}, Robust-xID improved upon Weighted and Soft-xID.

\section{Comparison to prior work}
We compare Robust-xID to prior work in self-supervised learning.
We train our models on the Kinetics dataset, using an 18-layer R(2+1)D model~\cite{tran2018closer} for the video, and a 9-layer 2D ConvNet with batch normalization for the audio.
Video clips of length 8-frames and $112\times{}112$ resolution are extracted at 16fps, and the same data augmentations from~\cref{sec:experiments} are used.
We extract audio clips of length 2s at 24KHz and compute log mel spectrograms with 128 time steps and 128 frequency bands.
All models are trained with the Adam optimizer with a batch size of 512 distributed across 8 12Gb GPUs. We warm-up the models for 200 epochs by training on the xID loss alone with a learning rate of $5e-4$.
The models are then trained with sample weights and cycle consistent soft targets for an additional 200 epochs using a cosine learning rate schedule from $5e\!-\!4$ to $5e\!-\!5$.

After pre-training, models are evaluated on UCF and HMDB. We fine-tune the models using either 8 or 32 frame clips for action recognition and report the top-1 accuracy of video level predictions (with 10 clips per video) in \cref{tab:sota}.
The proposed procedure outperformed all prior work where pretraining is limited to a single node (8 GPUs), and even outperformed methods like SeLaVi, which require 8$\times$ more compute for training.
We also conducted a close comparison to the CMA procedure of~\cite{avid} (xID+CMA). 
While CMA can also partially address the problem of faulty negatives, Robust-xID showed better performance. Robust-xID is also easier to implement as it identifies both faulty positives and negatives in a simpler online fashion.
We note that xID+CMA is a faithful implementation of AVID+CMA~\cite{avid}, as it follows the original code with improved data augmentations.
However, the results reported for xID+CMA are lower than those originally reported in~\cite{avid} because 1) distributed training was conducted on 8 GPUs instead of 64 (large batch sizes are known to have a substantial impact on contrastive learning performance~\cite{simclr, mocov2, swav}), and 2)~\cite{avid} is trained and evaluated with videos of higher resolution (224 instead of 112). By training the proposed model with a larger batch size, we expect the performance to improve further.

\begin{table}[t!]
\centering
\caption{{\bf Comparison to prior work (finetuning).} Performance on the downstream UCF and HMDB datasets by full network fine-tuning after pre-training on Kinetics. We report top-1 accuracy of video level predictions (10 clips per video). We also list the video encoder, amount of compute used for pre-training and the fine-tuning resolution.}
\label{tab:sota}
\resizebox{\linewidth}{!}{%
\definecolor{rc1}{RGB}{235,235,235}
\definecolor{rc2}{RGB}{255,255,255}
\begin{tabular}{lccccc}
\toprule
\bf Method & \bf Model & \bf \thead{Compute\\\# GPUs} & \bf \thead{Finetuning\\Resolution} & \bf UCF & \bf HMDB \\ \midrule\midrule
\rowcolor{rc1}
DPC~\cite{dpc}  & S3D        & 4 & $25\!\times\!128^2$ & 75.7 & 35.7 \\
\rowcolor{rc2}
CBT~\cite{cbt}  & S3D        & 8 & $16\!\times\!112^2$ & 79.5 & 44.6 \\
\rowcolor{rc1}
Multisensory~\cite{multisensory}  
                & 3D-ResNet18 & 3 & $32\!\times\!224^2$ & 82.1 & -- \\
\rowcolor{rc2}
AVTS~\cite{bruno_avts}  
                & MC3-18 & 4 & $25\!\times\!224^2$ & 84.1 & 52.5 \\
\rowcolor{rc1}
SeLaVi~\cite{selavi}
                & R(2+1)D-18 & 64 & $32\!\times\!112^2$ & 83.1${}^*$ & 47.1${}^*$ \\
\rowcolor{rc2}
                & R(2+1)D-18 & 64 & $8\!\times\!224^2$ & 74.2${}^*$ & 39.0${}^*$ \\
\rowcolor{rc2}
\multirow{-2}{*}{XDC~\cite{xdc}} 
                & R(2+1)D-18 & 64 & $32\!\times\!224^2$ & 86.8${}^*$ & 52.6${}^*$ \\
\rowcolor{rc1}
                & R(2+1)D-18 & 64 & $8\!\times\!224^2$ & 83.7${}^*$ & 49.5${}^*$ \\
\rowcolor{rc1}
\multirow{-2}{*}{AVID-CMA~\cite{avid}} 
                & R(2+1)D-18 & 64 & $32\!\times\!224^2$ & 87.5${}^*$ & 60.8${}^*$ \\
\rowcolor{rc2}
GDT~\cite{gdt}  & R(2+1)D-18 & 64 & $32\!\times\!224^2$ & 89.3${}^*$ & 60.0${}^*$ \\
% \rowcolor{rc1}
%                 & R(2+1)D-18 & 8 & $8\!\times\!112^2$ & 79.8 & 48.4 \\
% \rowcolor{rc1}
% \multirow{-2}{*}{xID} 
%                 & R(2+1)D-18 & 8 & $32\!\times\!112^2$ & 84.3 & 52.6 \\
\rowcolor{rc1}
                & R(2+1)D-18 & 8 & $8\!\times\!112^2$ & 80.6 & 48.6 \\
\rowcolor{rc1}
\multirow{-2}{*}{xID+CMA~\cite{avid}}
                & R(2+1)D-18 & 8 & $32\!\times\!112^2$ & 84.9 & 54.7 \\
\hline
\rowcolor{rc2}
                & R(2+1)D-18 & 8 & $8\!\times\!112^2$ & 81.9 & 49.5 \\
\rowcolor{rc2}
\multirow{-2}{*}{Robust-xID}
                & R(2+1)D-18 & 8 & $32\!\times\!112^2$ & \bf 85.6 & \bf 55.0 \\ 
\bottomrule
\multicolumn{6}{l}{\footnotesize \it ${}^*$ Models pre-trained with more than one compute node (8 GPUs).}
\end{tabular}%
}
\vspace{-8pt}
\end{table}

\begin{table}[t!]
\centering
\caption{{\bf Retrieval performance} on UCF and HMDB datasets after pre-training on Kinetics for different numbers of retried neighbors.}
\label{tab:sota-retrieval}
\resizebox{0.85\linewidth}{!}{%
\definecolor{rc1}{RGB}{235,235,235}
\definecolor{rc2}{RGB}{255,255,255}
\begin{tabular}{lcccccccc}
\toprule
\multirow{2}{*}{\bf Method} & \multicolumn{3}{c}{\bf UCF} && \multicolumn{3}{c}{\bf HMDB} \\ \cmidrule{2-4}\cmidrule{6-8}
& R@1 & R@5 & R@20 & & R@1 & R@5 & R@20 \\ \midrule\midrule
% \rowcolor{rc1}
% 3D-Puzzle~\cite{3d_puzzle}  & 19.7 & 28.5 & 40.0 && - & - & - \\
% \rowcolor{rc2}
% OPN~\cite{opn}              & 19.9 & 28.7 & 40.6 && - & - & - \\
% \rowcolor{rc1}
% ST Order~\cite{storder}     & 25.7 & 36.2 & 49.2 && - & - & - \\
\rowcolor{rc1}
SpeedNet~\cite{speednet}    & 13.0 & 28.1 & 49.5 && - & - & - \\
% \rowcolor{rc1}
% Clip Order~\cite{cliporder} & 14.1 & 30.3 & 51.1 && 7.6 & 22.9 & 48.8 \\
\rowcolor{rc2}
VCP~\cite{vcp}              & 18.6 & 33.6 & 53.5 && 7.6 & 24.4 & 53.6 \\
\rowcolor{rc1}
VSP~\cite{vsp}              & 24.6 & 41.9 & 76.9 && 10.3 & 26.6 & 54.6 \\
\rowcolor{rc2}
CoCLR~\cite{CoCLR}          & 55.9 & 70.8 & 82.5 && 26.1 & 45.8 & 69.7 \\
\rowcolor{rc1}
SeLaVi~\cite{selavi}        & 52.0 & 68.6 & 84.5 && 24.8 & 47.6 & 75.5 \\
\rowcolor{rc2}
GDT~\cite{gdt}              & 57.4 & 73.4 & 88.1 && 25.4 & 51.4 & 75.0 \\
% \rowcolor{rc1}
% xID                         & 57.9 & 75.2 & 88.9 && 28.4 & 52.2 & 76.9 \\
\rowcolor{rc1}
xID+CMA~\cite{avid}         & 60.1 & 76.6 & 90.1 && 29.7 & 54.4 & 77.1 \\
\hline
\rowcolor{rc2}
Robust-xID                  & \bf 60.9 & \bf 79.4 & \bf 90.8 && \bf 30.8 & \bf 55.8 & \bf 79.7 \\
\bottomrule
\end{tabular}%
}
\end{table}

\begin{table}[t!]
\centering
\caption{{\bf Few-shot learning} on UCF and HMDB after pre-training on Kinetics. Classification is conducted using a one-vs-all SVM trained on the pretrained features of $n$ images per class. Top-1 accuracy is reported for $n\in\{1, 5, 20\}$.}
\label{tab:sota-fewshot}
\resizebox{0.9\linewidth}{!}{%
\definecolor{rc1}{RGB}{235,235,235}
\definecolor{rc2}{RGB}{255,255,255}
\begin{tabular}{lcccccccc}
\toprule
\multirow{2}{*}{\bf Method} & \multicolumn{3}{c}{\bf UCF} && \multicolumn{3}{c}{\bf HMDB} \\ \cmidrule{2-4}\cmidrule{6-8}
& 1-shot & 5-shot & 20-shot & & 1-shot & 5-shot & 20-shot \\ \midrule\midrule
\rowcolor{rc1}
3D-RotNet~\cite{3drotnet}    & 15.0 & 31.5 & 47.1 && - & - & - \\
\rowcolor{rc2}
GDT~\cite{gdt}              & 26.3 & 42.4 & 49.4 && 13.4 & 15.6 & 20.8 \\
% \rowcolor{rc1}
% xID                         & 28.7 & 51.0 & 65.4 && 13.3 & 23.6 & 33.1 \\
\rowcolor{rc1}
xID+CMA~\cite{avid}         & 30.8 & 53.1 & 66.9 && 13.5 & 25.0 & 33.6 \\
\hline
\rowcolor{rc2}
Robust-xID                  & \bf 32.8 & \bf 54.6 & \bf 67.8 && \bf 14.1 & \bf 25.9 & \bf 34.9 \\
\bottomrule
\end{tabular}%
}
\end{table}

We also compare the learned representations to prior work without fine-tuning. Following~\cite{selavi,gdt}, we conducted retrieval experiments, and report the retrieval performance $R@K$ for $K=1$, $K=5$ and $K=20$ neighbors in \cref{tab:sota-retrieval}. The retrieval protocol was described in \cref{sec:experiments}.
Following \cite{3drotnet,gdt}, we also assessed the few-shot learning performance of Robust-xID models on UCF and HMDB. For the few-shot evaluation, we average the pretrained max-pooling features of 10 clips per video. The features from $n$ videos per class are then used to learn a one-vs-all linear SVM classifier with $C=1$. We report the top-1 accuracy averaged over 50 trials in \cref{tab:sota-fewshot}.
On both the retrieval and few-shot learning tasks, Robust-xID improves significantly over all prior work, reaffirming the importance of mitigating the training noise introduced by faulty positives and faulty negatives.

\section{Discussion and future work}
We identified and tackled two significant sources of noisy training signals in audio-visual instance discrimination, namely instances with weak audio-visual correspondence (or faulty positives) and semantically similar negatives (or faulty negatives). 
We showed the impact of faulty correspondences on representation learning by removing them using an oracle with access to ground-truth annotations.
We then proposed a method that mitigates the impact of faulty correspondences without relying on ground-truth annotations.
Extensive analysis and experimental evaluations show that the proposed procedure enhances representation learning and improves transfer performance significantly.

Our findings show that cross-modal learning should be seen as a problem of learning with noisy targets. While we propose two specific methods to address faulty positives and faulty negatives (\ie weighting and soft targets), there is a rich literature regarding supervised learning with noisy labels. Developing methods that tackle noisy correspondences are a promising avenue for future research. Furthermore, we focused on audio-visual learning, but other pairs of modalities such as RGB and flow or text from instructional videos also present similar problems. We believe that our method will also benefit cross-modal learning from other modalities.

\vspace{8pt}
\paragraph{Acknowledgements}
This work was partially funded by NSF awards IIS-1924937, IIS-2041009, and gifts from Amazon and Qualcomm. We also acknowledge and thank the use of the Nautilus platform for some of the experiments above.

{\small
\bibliographystyle{ieee_fullname}
\bibliography{refs}
}

\vfill\pagebreak\ \vfill\pagebreak
\appendix
\section{Parametric studies}
We provide a parametric study of key Robust-xID hyper-parameters.

\begin{figure}[b]
    \centering
    \includegraphics[width=\linewidth]{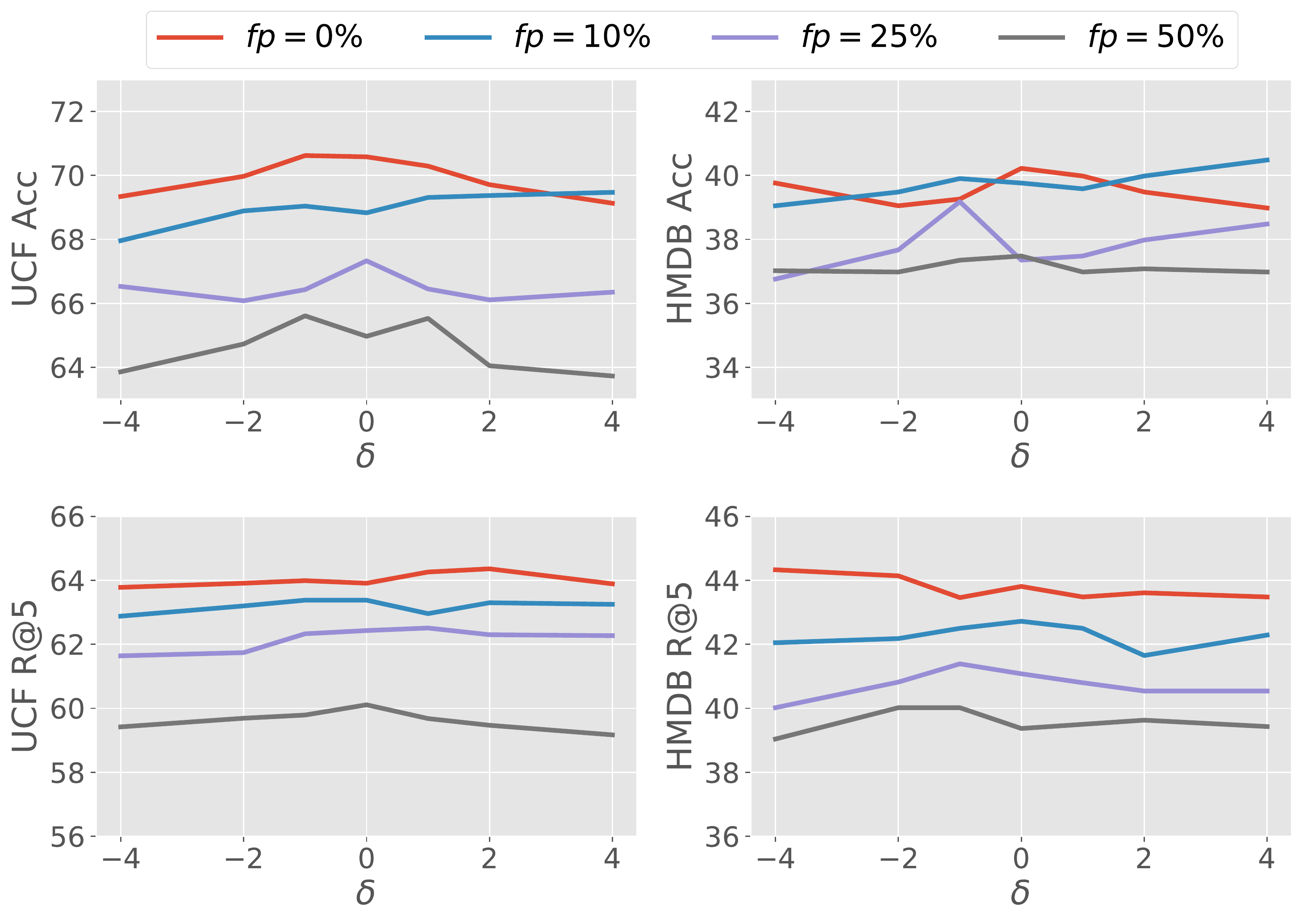}
    \caption{{\bf Effect of shape parameter $\delta$ in Weighted-xID}. Transfer learning performance is evaluated on two datasets (UCF and HMDB) under two protocols (full finetuning and retrieval). For the fine-tuning protocol, we report final accuracy of video level predictions. For the retrieval protocol, we report $R@5$.}
    \label{fig:delta}
\end{figure}

\paragraph{Weight function shape parameter $\delta$}
One critical parameter of Weighted-xID is the shape parameter $\delta$, which specifies the mid-point location of the weight function. For example, when $\delta=-2$, the midpoint is located at $\mu-2\sigma$ where $\mu$ and $\sigma$ are the sample mean and standard deviation of the scores $\vbar_i^T\abar_i$. This means that for $\delta=-2$, the majority of samples will have a weight of $1$, and only a small fraction will have a weight close to $w_{\min}$. As $\delta$ increases, the proportion of samples that are down-weighted also increases. 
To study the impact of $\delta$, we trained several models using Weighted-xID with different values of $\delta$ and for different amounts of injected faulty positives $n_0$. Other hyper-parameters were kept at their default values $w_{\min}=0.25$ and $\kappa=0.5$. The transfer performance is shown in~\cref{fig:delta}.
As can be seen, the proposed robust xID procedure is not very sensitive to this hyper-parameter. This suggests that Robust-xID can help representation learning as long as clear faulty positives are suppressed.

\begin{figure}[b]
    \centering
    \includegraphics[width=\linewidth]{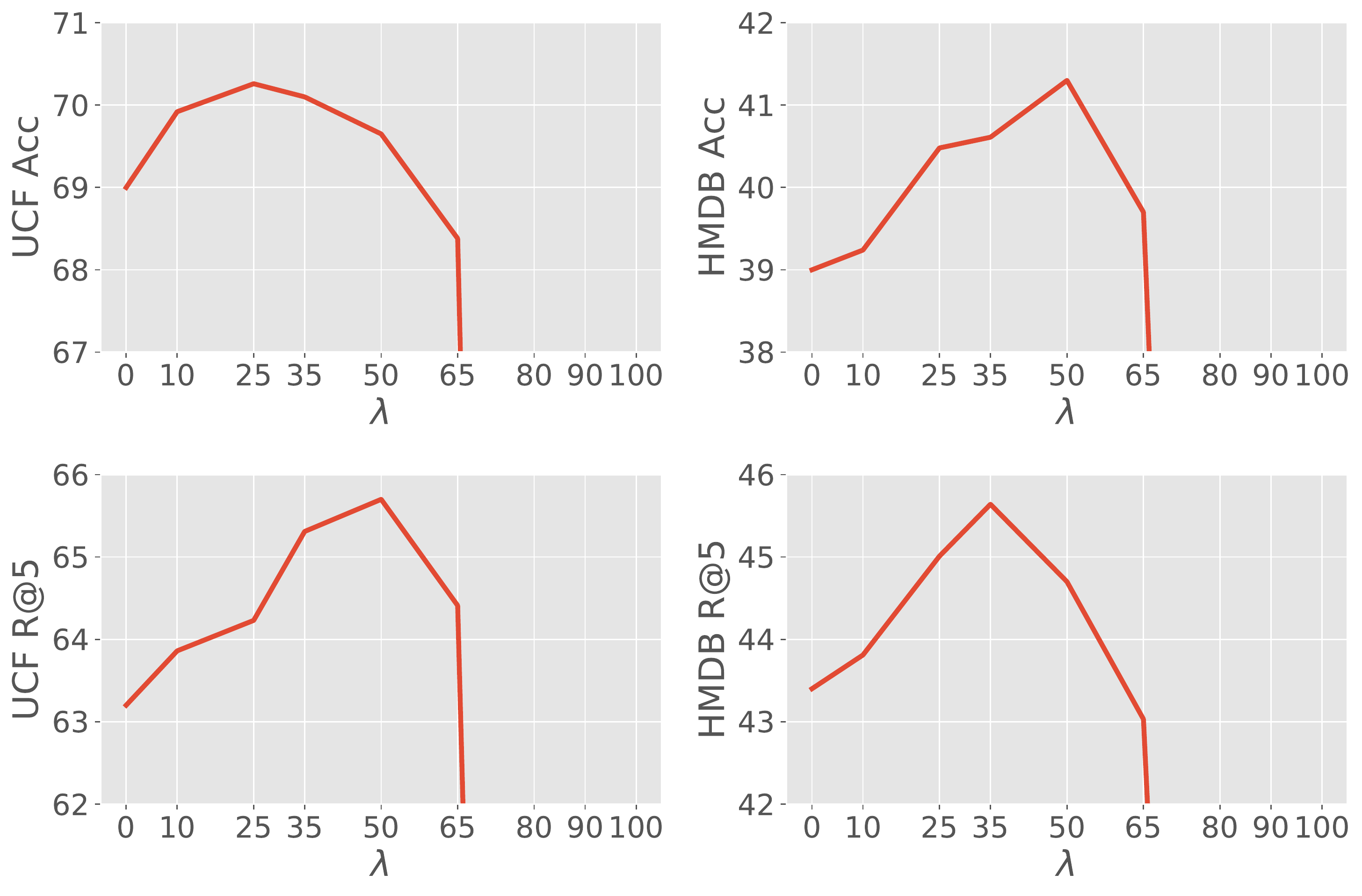}
    \caption{{\bf Effect of mixing coefficient $\lambda$ in Soft-xID.} Transfer learning performance is evaluated on two datasets (UCF and HMDB) under two protocols (full finetuning and retrieval). For the fine-tuning protocol, we report final accuracy of video level predictions. For the retrieval protocol, we report $R@5$.}
    \label{fig:lambda}
\end{figure}

\begin{figure*}
    \centering
    \begin{tabular}{c}
        \includegraphics[width=\linewidth]{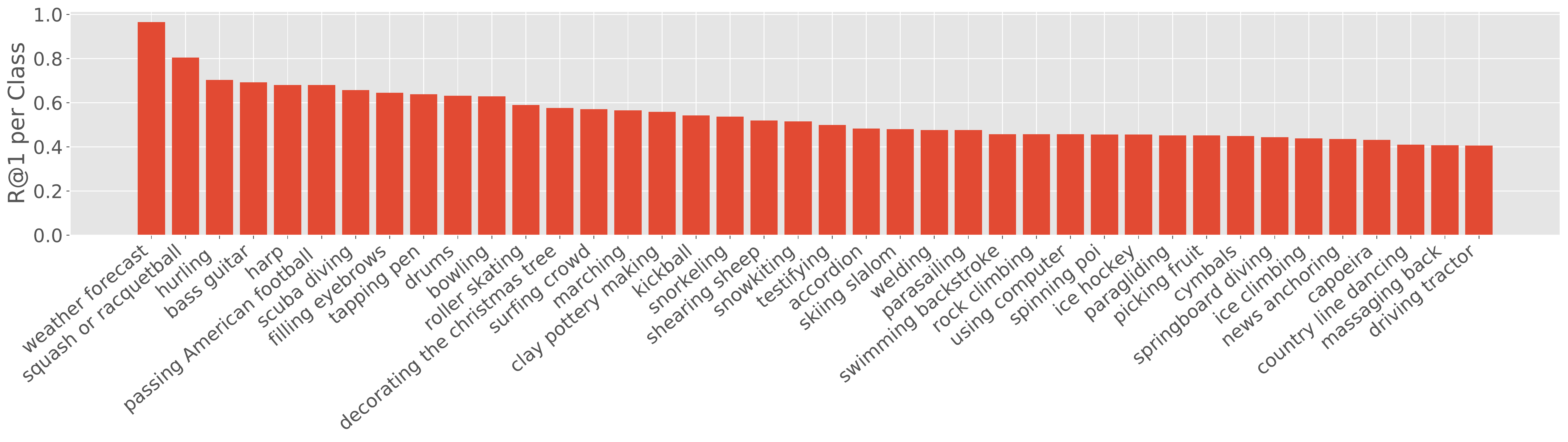} \\
        \includegraphics[width=\linewidth]{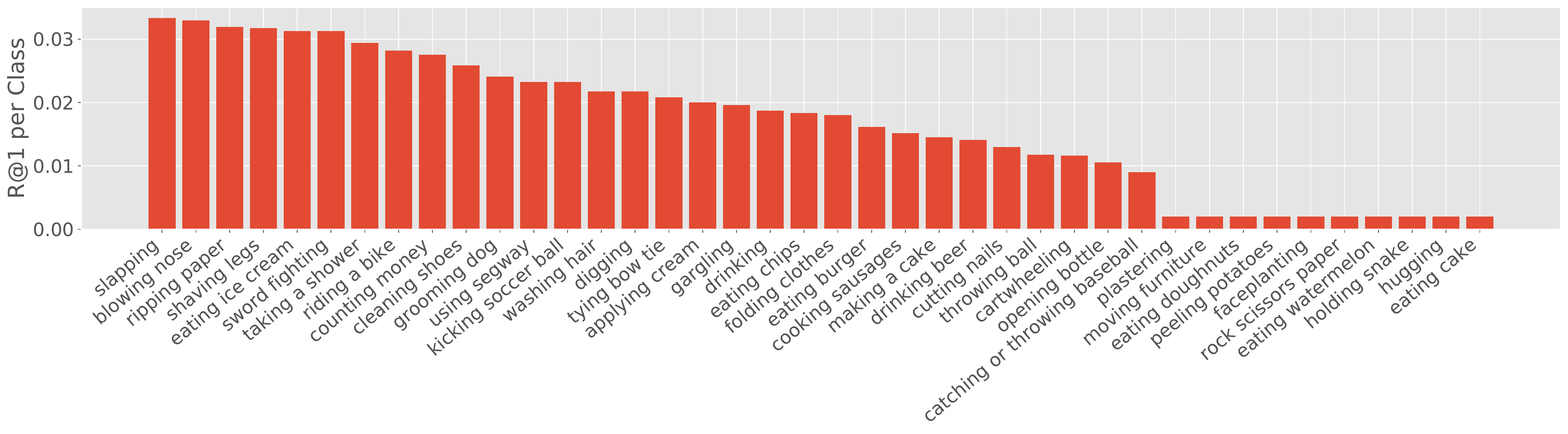}
    \end{tabular}
    \caption{{\bf Best and worse Kinetics classes.} For each class, we depict the top-1 retrieval performance ($R@1$) averaged across all images of each class. The plot above shows the top 40 classes and the plot below the bottom 40 classes. }
    \label{fig:top-bottom-classes}
\end{figure*}

\paragraph{Soft-xID: Mixing coefficient}
The mixing coefficient $\lambda$ specifies the degree to which the one-hot targets of instance discrimination are softened in Soft-xID. The one-hot instance discrimination targets are used when $\lambda=0$. As $\lambda$ increases, the softening scores $S(j|i)$ are increasingly used to adjust the one-hot targets.
To study the impact of the mixing coefficient $\lambda$, we trained several models using Soft-xID with various values of $\lambda$. Cycle consistent targets were used as the softening strategy. \cref{fig:lambda} shows the transfer performance of the learned models on UCF and HMDB under the fine-tuning and retrieval protocols. The trend is consistent across the two datasets and two evaluation protocols. Softening the instance discrimination targets enhances representation learning, with the optimal performance achieved with a mixing coefficient between $0.25$ and $0.5$. However, as the mixing coefficient increases substantially $\lambda>0.65$, the targets are derived from the model prediction alone and disregard instance labels. In this case of large $\lambda$, the pre-training fails completely, \ie, the learned representations have very low transfer performance.

\section{Additional analysis}
The proposed approach learns high-quality feature representations that can be used to discriminate several action classes. This was shown in the main paper by reporting transfer learning results. We now provide additional qualitative evidence and analysis.

\begin{figure*}[t!]
    \centering
    \begin{tabular}{c|c}
        \includegraphics[width=0.47\linewidth]{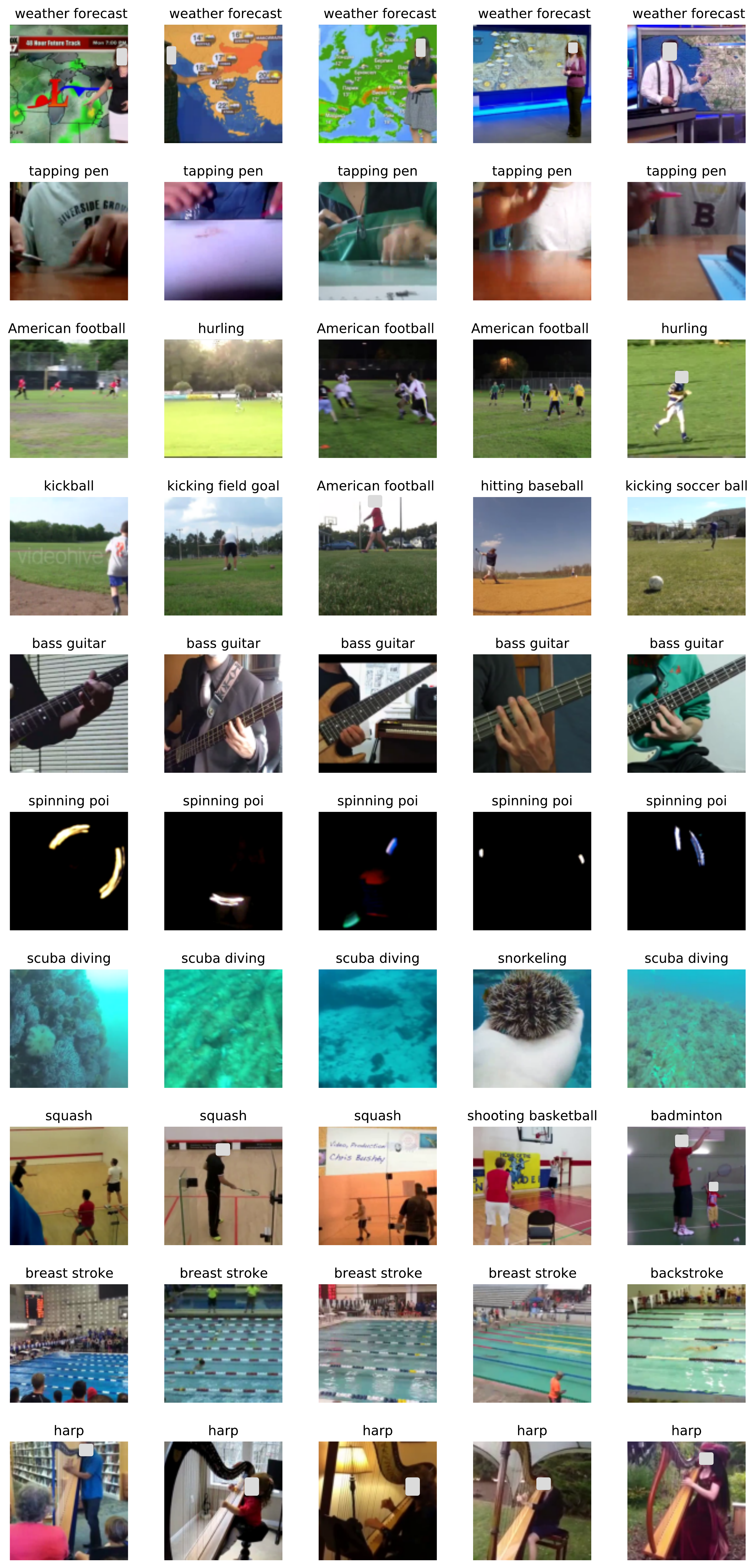} &  
        \includegraphics[width=0.47\linewidth]{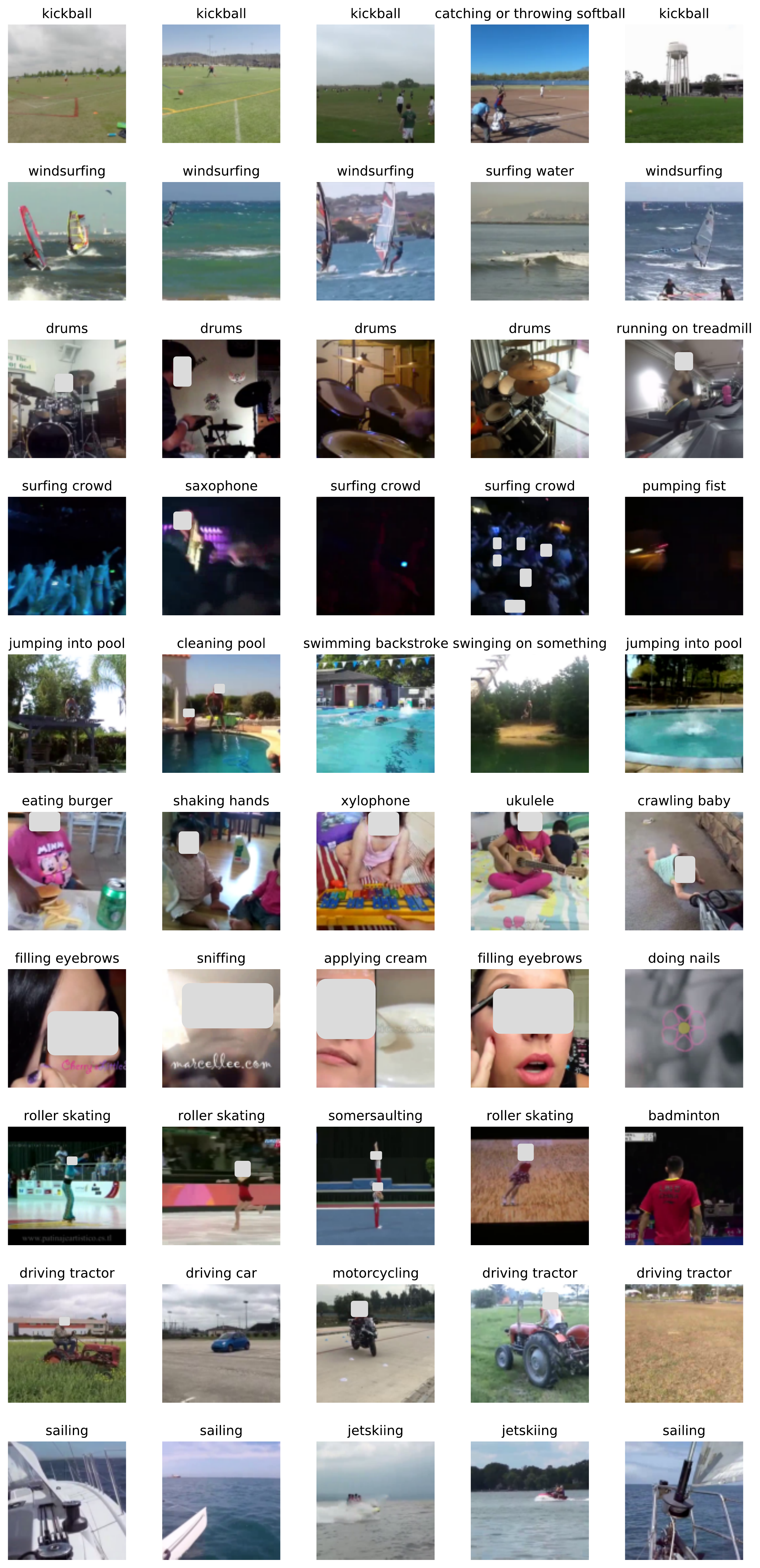}
    \end{tabular}
    \caption{{\bf Retrievals.} In each row, the first image depicts the query video, and the following four images depict the top 4 retrievals. The corresponding Kinetics class description is provided above each frame.  Each video is represented by a $4\times4\times512$ feature map produced by the video encoder learned using Robust-xID on the full Kinetics dataset. Euclidean distance is used to determine video similarity.}
    \label{fig:retrievals}
\end{figure*}

\paragraph{Retrieval}
For each video, we extracted $4\times4\times512$ feature maps from the video encoder learned using Robust-xID on the full Kinetics dataset. \cref{fig:retrievals} depicts the top 4 closest videos for several query samples.
As can be seen, Robust-xID produces highly semantic features, enabling correct retrievals for a large number of videos spanning a large number of classes. Furthermore, even when a video of a different class is retrieved, the errors are intuitive (for example, the confusion between `American football` and `Hurling` in the third row). Failure cases also seem to be correlated with classes that are hard to distinguish from the audio alone (eg, different types of kicking sports or swimming strokes).

\paragraph{Class-based analysis}
To better understand which classes are better modeled by the Robust-xID framework, we measured the top-1 retrieval performance ($R@1$) averaged across all images of each class. Similar to the analysis above, each video is represented by a $4\times4\times512$ feature map extracted from a video encoder learned using Robust-xID on the full Kinetics dataset. \cref{fig:top-bottom-classes} depicts a list of Kinetics classes sorted by their average $R@1$ score. As can be seen, action classes which are often accompanied by long and distinctive sounds (\eg, squash, harp, drums, accordion, or scuba diving) tend to be more easily distinguished from others. In contrast, classes with less distinctive audio (\eg, making a cake, eating cake, or hugging) or classes where distinctive sounds are short-lived (\eg, blowing nose, gargling or kicking ball) are harder to model using a cross-modal audio-visual framework. As a result, the features learned for such classes are less discriminative.

\paragraph{Faulty positive detection performance}
To obtain a rough estimate of performance of the faulty positive detection procedure, we randomly sampled 100 videos from the 10000 most likely faulty positives, as identified by Robust-xID trained on the full Kinetics dataset. We then manually labeled them according to how related their audio and visual signals are. From those, 67 were clear faulty pairs; 24 contained narrative voice-overs (\ie, required natural language understanding to link the two modalities); and 9 samples were clearly misidentified.

\end{document}